\documentclass[runningheads]{llncs}
\usepackage[T1]{fontenc}

\usepackage{graphicx}

\usepackage{marvosym}

\usepackage{amssymb}

\usepackage{amsmath}

\usepackage{multirow}

\usepackage{booktabs}

\usepackage{threeparttable}

\usepackage[linesnumbered,ruled,vlined]{algorithm2e}

\begin{document}
\title{QORH: Learning Query-Specific Ordered Rule Hypergraphs for Inductive Knowledge Graph Link Prediction}
\titlerunning{QORH: Query-Specific Ordered Rule Hypergraphs}

\author{Naixing Xu\orcidID{0009-0005-2094-330X}}

\authorrunning{N. Xu et al.}

\institute{School of Software, Shandong University, Jinan, China \\
\email{naixingxu@mail.sdu.edu.com}
}

\maketitle    
\begin{abstract}
Knowledge graph link prediction involving unseen entities and relations requires learning transferable reasoning patterns from observed graph structure. Existing methods based on relation graphs primarily capture pairwise dependencies, limiting their ability to preserve the compositional semantics, relation ordering, and role information of complete path rules. To address this limitation, we propose QORH (Query-specific Ordered Rule Hypergraph), which jointly encodes rule heads and ordered rule bodies as hyperedges conditioned on the query relation, explicitly modeling joint dependencies among multiple relations. QORH performs learnable rule selection using statistical evidence, such as rule confidence and coverage, and enhances relation representations through role-aware hypergraph message passing to guide entity-level reasoning. The model’s shared parameters are independent of specific entity or relation identifiers, enabling inductive prediction for unseen entities and relations. Extensive experiments on entity-inductive and fully inductive benchmarks demonstrate that QORH significantly outperforms the state-of-the-art baselines included in our evaluation, validating the effectiveness of rule hypergraph modeling for inductive link prediction.

\keywords{Knowledge graph link prediction  \and Inductive reasoning \and Ordered rule hypergraphs \and Higher-order relational modeling \and Graph neural networks.}
\end{abstract}

\section{Introduction}

\begin{figure}[htbp]
  \centering
  \includegraphics[width=1.0\textwidth]{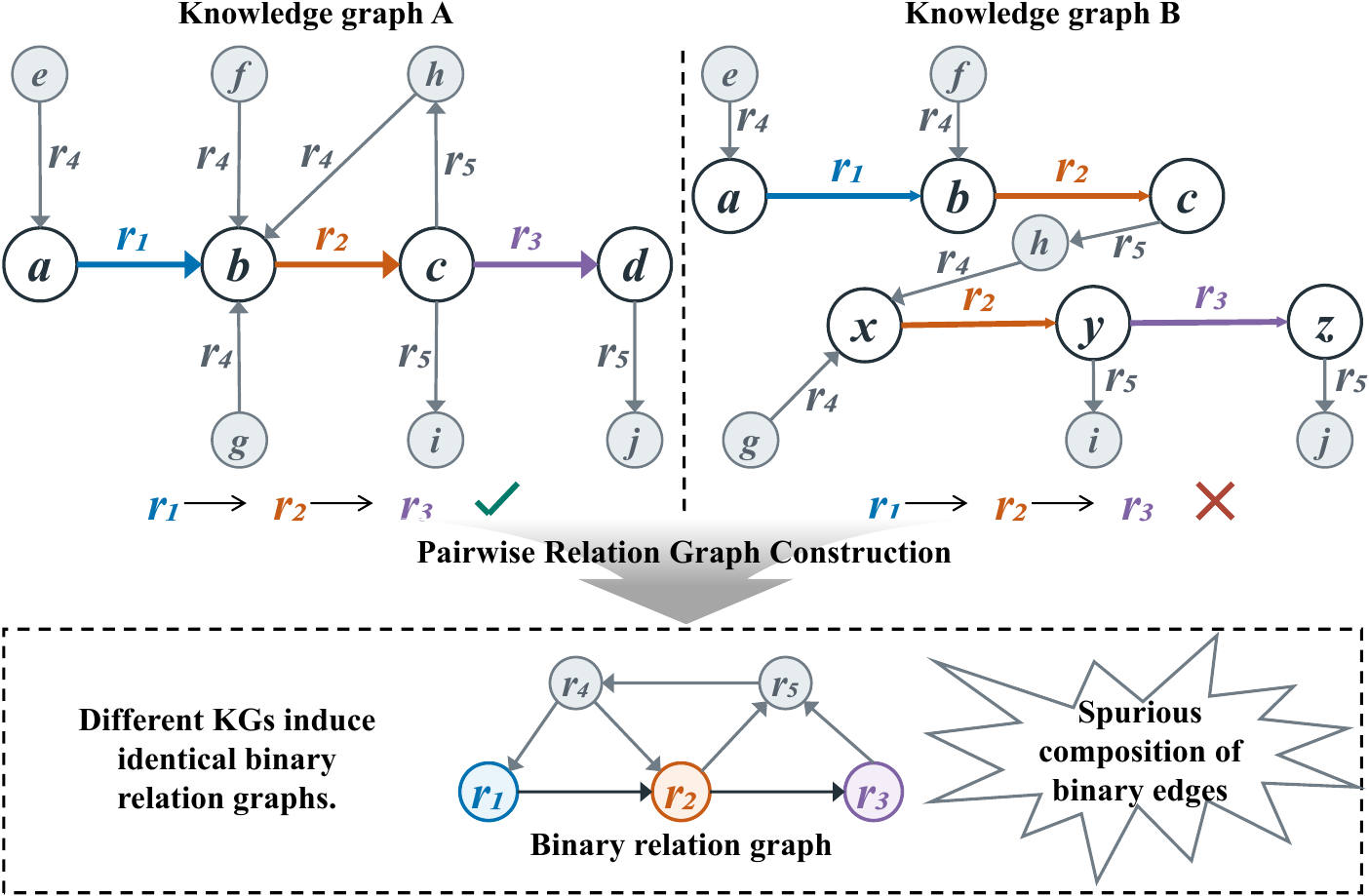}
  \caption{Spurious composition of binary edges in relation-level auxiliary graphs. The two KGs induce identical binary relation graphs, yet only KG A contains a continuous path with the relation sequence $(r_1,r_2,r_3)$.}
  \label{fig:problem_to_solve_spurious-composition}
\end{figure}

Knowledge graphs (KGs) represent entities and their semantic relationships as structured triples $(h,r,t)$, where $h$, $t$, and $r$ denote the head entity, tail entity, and their relation, respectively. Owing to their rich semantics and structured representation capabilities, KGs have been widely applied to recommender systems~\cite{wangRippleNetPropagatingUser2018,wangKGATKnowledgeGraph2019}, question answering~\cite{bordesQuestionAnsweringSubgraph2014,yihSemanticParsingStaged2015}, biomedicine~\cite{belleauBio2RDFMashupBuild2008,himmelsteinSystematicIntegrationBiomedical2017}, and other domains~\cite{zhangERNIEEnhancedLanguage2019,liuKBERTEnablingLanguage2020}.

However, KGs are typically incomplete because of the high cost of knowledge acquisition and the continuous evolution of real-world knowledge. Knowledge graph link prediction aims to infer missing facts from observed ones and has become a central task in KG research. Although conventional embedding methods such as TransE~\cite{bordesTranslatingEmbeddingsModeling2013}, DistMult~\cite{yangEmbeddingEntitiesRelations2014}, ComplEx~\cite{trouillonComplexEmbeddingsSimple2016}, and RotatE~\cite{sunRotatEKnowledgeGraph2018} have achieved notable progress, they generally follow the transductive setting, assuming that all entities and relations encountered during inference have been observed during training. Consequently, they struggle to generalize to unseen entities and relations.

To handle unseen entities, entity-inductive methods such as GraIL~\cite{teruInductiveRelationPrediction2020}, TACT~\cite{chenTopologyAwareCorrelationsRelations2021}, MorsE~\cite{chenMetaKnowledgeTransferInductive2022}, and Content Pooling~\cite{suContextPoolingQueryspecific2025} learn transferable local structures and relational patterns without relying on entity-specific representations. Nevertheless, these methods typically assume a shared relation set between the training and test graphs, limiting their applicability to open-world scenarios in which both entities and relations continuously emerge.

Accordingly, fully inductive KG link prediction requires models to generalize simultaneously to unseen entities and relations. Methods such as RMPI~\cite{gengRelationalMessagePassing2023}, InGram~\cite{leeINGRAMInductiveKnowledge2023}, and TARGI~\cite{dingGlobalTopologyRelationGraph2025} enable fully inductive reasoning by learning transferable relational structures. KG foundation models such as ULTRA~\cite{galkinFoundationModelsKnowledge2024} and GraphOracle~\cite{duGraphOracleEfficientFullyInductive2026a} further learn general reasoning patterns independent of specific entity and relation vocabularies through pretraining across multiple KGs, achieving strong performance on various inductive benchmarks with limited fine-tuning on target KGs.

Despite the progress made by fully inductive KG link prediction methods such as InGram~\cite{leeINGRAMInductiveKnowledge2023}, ULTRA~\cite{galkinFoundationModelsKnowledge2024}, TARGI~\cite{dingGlobalTopologyRelationGraph2025}, and GraphOracle~\cite{duGraphOracleEfficientFullyInductive2026a}, their relation modeling is generally limited to pairwise dependencies induced by shared entities, making it difficult to preserve the compositional semantics, relation roles, and overall reliability of ordered multi-relation rules. As illustrated in Figure~\ref{fig:problem_to_solve_spurious-composition}, given a KG $\mathcal{G}=(\mathcal{V},\mathcal{R},\mathcal{F})$, GraphOracle~\cite{duGraphOracleEfficientFullyInductive2026a} establishes a directed edge in its relation-level auxiliary graph whenever two relations are sequentially connected through an intermediate entity: $(r_i,r_j)\in\mathcal{E}_R \iff \exists\,u,x,v\in\mathcal{V}: (u,r_i,x)\in\mathcal{F}\land(x,r_j,v)\in\mathcal{F},$ where $\mathcal{V}$, $\mathcal{R}$, and $\mathcal{F}$ denote the entity set, relation set, and factual triple set, respectively, and $\mathcal{E}_R$ is the directed edge set of the relation-level auxiliary graph. This construction retains only local adjacency between relations and cannot determine whether multiple pairwise dependencies originate from the same complete path. For example, both knowledge graph A and knowledge graph B contain $r_1\rightarrow r_2$ and $r_2\rightarrow r_3$, whereas only knowledge graph A contains the complete sequence $r_1\rightarrow r_2\rightarrow r_3$. In knowledge graph B, the two relation pairs arise from different paths and therefore cannot form a valid reasoning chain. We refer to this limitation as the “spurious composition of binary edges,” indicating that pairwise dependencies alone are insufficient to recover higher-order joint structures among multiple relations.

Moreover, as discussed in GraphOracle, the performance of existing KG foundation models often depends substantially on fine-tuning over target datasets~\cite{duGraphOracleEfficientFullyInductive2026a}. Our data analysis in Section~\ref{sec:Baselines and Experimental Setup} further reveals overlaps between some pretraining data and inductive test sets, exposing models to information that should remain unseen during evaluation and thereby causing data leakage. Consequently, the resulting performance may conflate pretraining memorization with inductive generalization, making it difficult to assess the models’ true capabilities under strict inductive protocols. Therefore, achieving reliable link prediction for unseen entities and even unseen relations under strict inductive evaluation protocols remains an important open problem. 

To address these limitations, we propose a query-specific rule hypergraph enhancement method that represents ordered path rules as hyperedges connecting rule-head and rule-body relations, thereby preserving their compositional semantics, ordering, and role information. A lightweight multilayer perceptron jointly considers rule statistics, such as confidence and coverage, to select high-quality hyperedges, while role-aware relation–hyperedge–relation message passing enhances relation representations for entity-level reasoning and link prediction. Following the entity-inductive protocol of GraIL~\cite{teruInductiveRelationPrediction2020} and the fully inductive protocol of InGram~\cite{leeINGRAMInductiveKnowledge2023}, our method achieves state-of-the-art performance in both inductive settings. The main contributions of this work are summarized as follows:

\begin{itemize}
    \item We propose a query-specific rule hypergraph enhancement method that explicitly models ordered multi-relation rules through learnable rule selection and role-aware hypergraph propagation.
    \item We identify the loss of multi-relation rule information and the spurious composition of binary edges caused by binary relation-level auxiliary graphs, and analyze this limitation from the perspective of higher-order relational modeling.
    \item We conduct systematic experiments on 12 entity-inductive and 12 fully inductive datasets. The proposed method consistently outperforms existing state-of-the-art approaches on the major evaluation metrics, demonstrating strong inductive generalization and robustness.
\end{itemize}

\section{Related Work}
\label{sec:Related Work}
This section systematically reviews recent advances in entity-inductive and fully-inductive knowledge graph link prediction and surveys related work on hypergraphs for knowledge graph representation learning and link prediction.

\begin{figure}[htbp]
  \centering
  \includegraphics[width=1.0\textwidth]{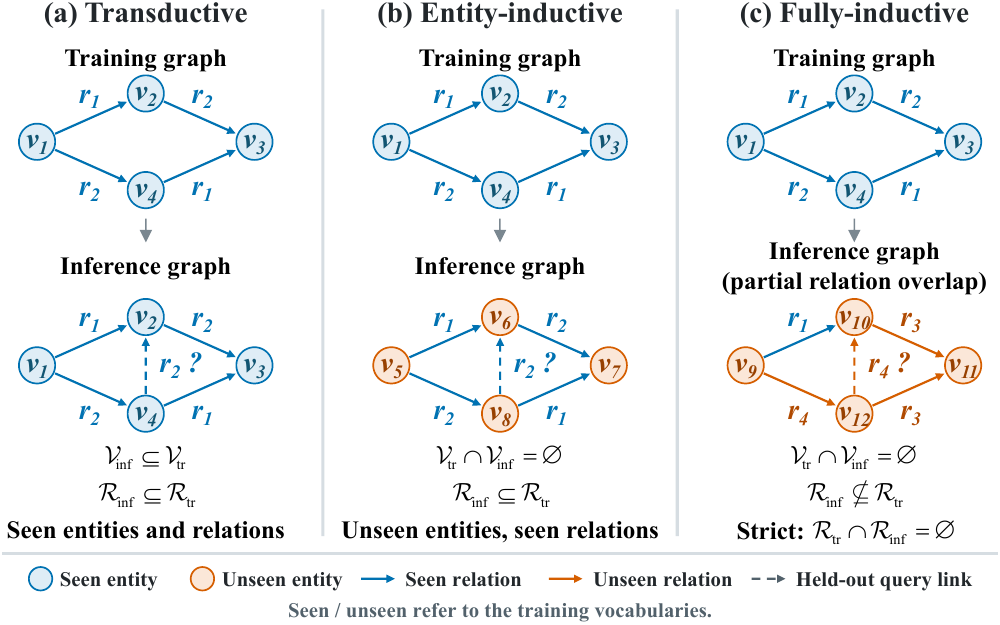}
  \caption{Illustration of knowledge graph link prediction settings. (a) Transductive: seen entities and relations. (b) Entity-inductive: unseen entities with seen relations. (c) Fully-inductive: unseen entities with partially or entirely unseen relations.}
  \label{fig:transductive_inductive_settings}
\end{figure}

\subsection{Entity-Inductive Knowledge Graph Link Prediction}

As illustrated in Figure~\ref{fig:transductive_inductive_settings}, let $\mathcal{V}_{\mathrm{tr}}$ and $\mathcal{V}_{\mathrm{inf}}$ denote the entity sets of the training and inference graphs, respectively, and let $\mathcal{R}_{\mathrm{tr}}$ and $\mathcal{R}_{\mathrm{inf}}$ denote their corresponding relation sets. Under the transductive setting, all entities and relations involved at inference time have already appeared during training; that is, $\mathcal{V}_{\mathrm{inf}}\subseteq\mathcal{V}_{\mathrm{tr}}$ and $\mathcal{R}_{\mathrm{inf}}\subseteq\mathcal{R}_{\mathrm{tr}}$. In contrast, the entity-inductive setting requires $\mathcal{V}_{\mathrm{tr}}\cap\mathcal{V}_{\mathrm{inf}}=\varnothing$ while retaining a shared relation vocabulary, thereby evaluating the model’s ability to generalize to unseen entities.

Logical rules are inherently independent of specific entities and therefore constituted an important early approach to entity-inductive reasoning. Neural LP~\cite{yangDifferentiableLearningLogical2017} represents relation compositions as differentiable matrix operations; DRUM~\cite{sadeghianDRUMEndToEndDifferentiable2019} improves the efficiency of learning long rules through recurrent rule modeling; and RNNLogic~\cite{quRNNLogicLearningLogic2021} treats logical rules as latent variables and jointly optimizes rule generation and reasoning.

More recently, graph neural networks have further advanced this field. GraIL~\cite{teruInductiveRelationPrediction2020} encodes the local enclosing subgraph surrounding each candidate triple to learn entity-independent structural patterns. CoMPILE~\cite{trouillonComplexEmbeddingsSimple2016} strengthens the interactions between entities and relations within directed subgraphs, thereby improving the modeling of asymmetric relations. TACT~\cite{chenTopologyAwareCorrelationsRelations2021} captures topological correlations between relation pairs to enrich the relational information contained in local subgraphs. MorsE~\cite{chenMetaKnowledgeTransferInductive2022} employs meta-learning to generate representations of unseen entities from their observed neighborhoods. Among path-based reasoning methods, NBFNet~\cite{zhuNeuralBellmanfordNetworks2021} formulates link prediction as a learnable generalized Bellman–Ford process; RED-GNN~\cite{zhangKnowledgeGraphReasoning2022} improves reasoning efficiency by sharing overlapping paths among candidate entities; and the recently proposed Context Pooling~\cite{suContextPoolingQueryspecific2025} selects logically relevant neighbors conditioned on a given query to suppress structural noise.

Although these methods can generalize to unseen entities, they typically assume that the training and inference graphs share the same relational semantics, allowing learned relation-specific parameters or rule templates to be reused during inference. Consequently, they have difficulty adapting to open knowledge graphs in which new relations continually emerge.

\subsection{Fully-Inductive Knowledge Graph Link Prediction}
\label{sec:Fully-Inductive Knowledge Graph Link Prediction}

As illustrated in Figure~\ref{fig:transductive_inductive_settings}(c), the fully-inductive setting further allows the inference graph to contain relations unseen during training, i.e., $\mathcal{R}_{\mathrm{inf}}\nsubseteq\mathcal{R}_{\mathrm{tr}}$. Under the strictest setting, $\mathcal{R}_{\mathrm{tr}}\cap\mathcal{R}_{\mathrm{inf}}=\varnothing$ also holds. Therefore, the model must derive representations of both unseen entities and unseen relations from the structure of the inference graph.

Existing methods primarily employ relation-level auxiliary graphs to learn transferable representations. RMPI~\cite{gengRelationalMessagePassing2023} performs relational message passing within local subgraphs; InGram~\cite{leeINGRAMInductiveKnowledge2023} constructs a weighted relation graph based on relation affinities and successively aggregates relation-level and entity-level information; and TARGI~\cite{dingGlobalTopologyRelationGraph2025} exploits global topological interactions to alleviate the biases introduced by local structures and relation frequencies.

In recent years, knowledge graph foundation models have demonstrated strong reasoning performance and generalization capabilities in inductive link prediction. ULTRA~\cite{galkinFoundationModelsKnowledge2024} learns transferable relative relation representations by categorizing the ways in which relations share entities into four interaction types: head-to-head, head-to-tail, tail-to-head, and tail-to-tail. The recently proposed GraphOracle~\cite{duGraphOracleEfficientFullyInductive2026a} constructs a sparse directed relation-dependency graph and generates relation representations through query-conditioned attention.

However, the relation-level auxiliary graphs employed by these methods are still predominantly composed of pairwise edges. Although such graphs can represent co-occurrence, topological, or logical dependencies between relations, they cannot reliably determine whether multiple pairwise dependencies originate from the same complete path. Consequently, the overall semantics of a rule, the positions of relations along the path, and their respective roles may be weakened. Furthermore, knowledge graph foundation models such as ULTRA~\cite{galkinFoundationModelsKnowledge2024} and GraphOracle~\cite{duGraphOracleEfficientFullyInductive2026a} require lengthy pretraining on multiple datasets, including WN18RR~\cite{dettmersConvolutional2DKnowledge2018} and FB15k-237~\cite{toutanovaObservedLatentFeatures2015}, whereas existing mainstream inductive link prediction benchmarks~\cite{teruInductiveRelationPrediction2020,leeINGRAMInductiveKnowledge2023} are constructed by partitioning WN18RR~\cite{dettmersConvolutional2DKnowledge2018} and FB15k-237~\cite{toutanovaObservedLatentFeatures2015}. Therefore, even when inductive evaluation is conducted in a zero-shot setting, i.e., without fine-tuning on the target knowledge graph, overlaps may still exist between the training data and the inductive test data (see Section~\ref{sec:Baselines and Experimental Setup}). As a result, the reported performance may not faithfully reflect the models’ generalization capabilities under strict inductive evaluation protocols.

Unlike the aforementioned methods, our approach strictly follows the inductive evaluation protocols introduced by GraIL~\cite{teruInductiveRelationPrediction2020} and InGram~\cite{leeINGRAMInductiveKnowledge2023}. For each query relation, it mines complete ordered path rules and represents each rule as a hyperedge connecting the rule head and multiple rule-body relations, thereby explicitly modeling multi-relation joint dependencies that cannot be fully captured by binary relation graphs.

\subsection{Hypergraphs for Knowledge Graph Representation and Link Prediction}

A hypergraph allows a single hyperedge to connect more than two nodes, enabling it to represent higher-order dependencies that ordinary graphs cannot fully capture. HGNN~\cite{fengHypergraphNeuralNetworks2019} learns higher-order data correlations through message passing and provides a general framework for subsequent research on hypergraph neural networks. Several studies have extended hypergraph modeling to more complex knowledge representation and knowledge graph link prediction scenarios. StarE~\cite{galkinMessagePassingHyperRelational2020} develops a message-passing mechanism for hyper-relational knowledge graphs with qualifiers, jointly encoding a primary triple and its auxiliary information. LHP~\cite{yangLHPLogicalHypergraph2023} represents logical conjunctions using directed hyperedges and treats logical dependencies among multiple nodes as an integrated structure for hyperlink prediction. ReAlE~\cite{fatemiKnowledgeHypergraphEmbedding2021} further improves the expressiveness of knowledge hypergraph representations from the perspective of relational algebra. Nevertheless, existing methods primarily focus on representing and completing native n-ary facts or qualified facts, and most operate under transductive settings in which the training and test stages share entity or relation vocabularies. By contrast, the application of hypergraphs to inductive knowledge graph link prediction remains at an early stage. In particular, how to construct transferable relation-level reasoning structures using hypergraphs in traditional knowledge graphs, where binary relations are represented as triples $(h,r,t)$, remains insufficiently explored.

\section{Methodology}
\label{sec:Methodology}
\begin{figure}[htbp]
  \centering
  \includegraphics[width=1.0\textwidth]{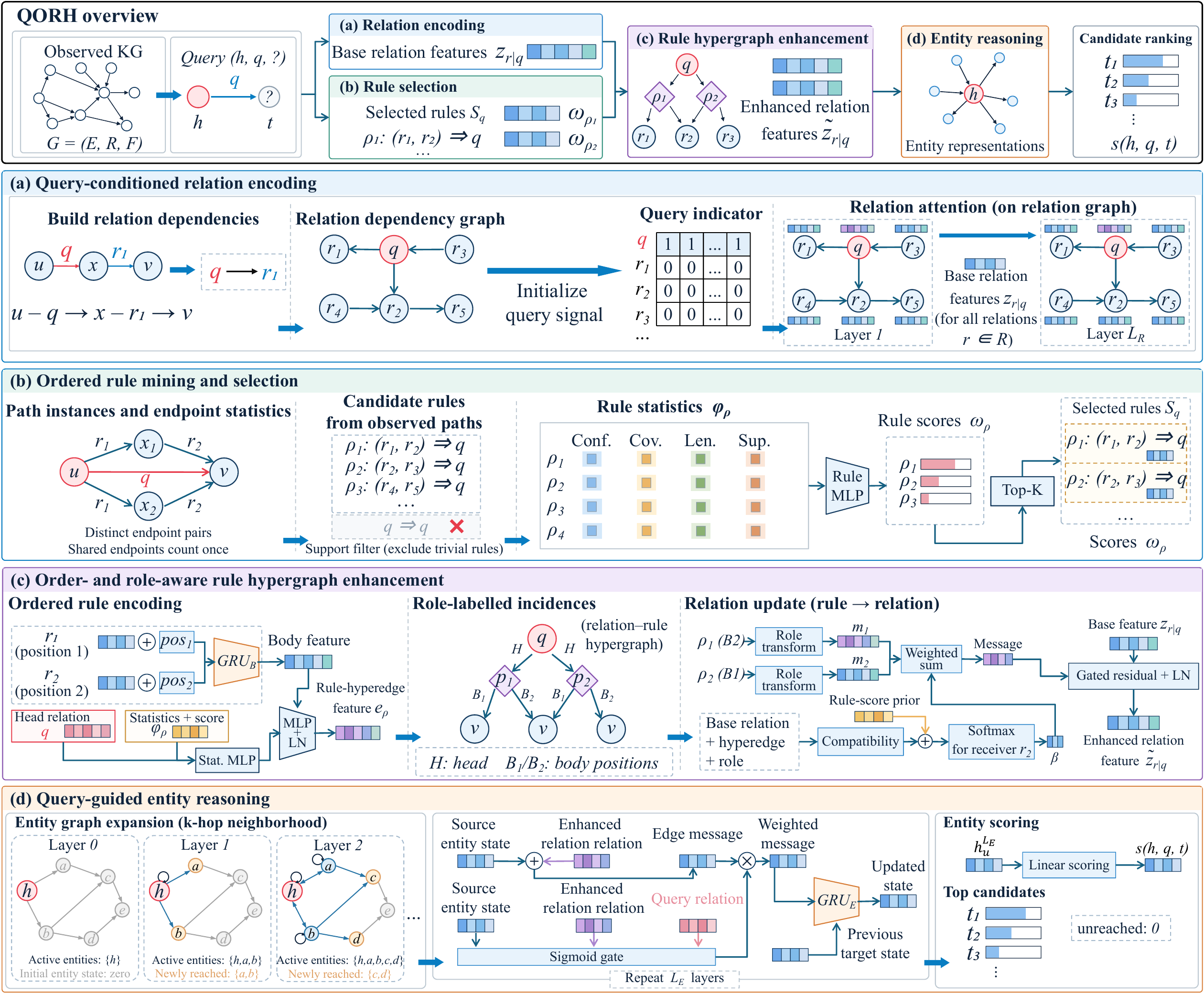}
  \caption{Overview of QORH: query-specific ordered rule hypergraphs for inductive knowledge graph link prediction.}
  \label{fig:methodology_overview}
\end{figure}

This section details the proposed Query-specific Ordered Rule Hypergraph (QORH) method, whose overall framework is illustrated in Figure~\ref{fig:methodology_overview}.

\subsection{Problem Formulation and Framework Overview}
\label{sec:Problem Formulation and Framework Overview}
Following the entity-inductive and fully inductive settings defined in Section~\ref{sec:Related Work}, let $\mathcal{G}_{\mathrm{obs}}=(\mathcal V,\mathcal R,\mathcal F_{\mathrm{obs}})$ denote the observed knowledge graph at the current stage, where $\mathcal V$, $\mathcal R$, and $\mathcal F_{\mathrm{obs}}$ are the sets of entities, relations, and observed facts available for reasoning, respectively. Given a head entity $h$ and a query relation $q$, the model ranks candidate tail entities $t\in\mathcal V$ for the query $(h,q,?)$ using the scoring function $s_\theta(h,q,t\mid\mathcal G_{\mathrm{obs}})$, where $\theta$ denotes the learnable parameters.

To unify head and tail entity prediction, we augment each observed fact $(u,r,v)$ with its inverse fact $(v,r^{-1},u)$, yielding the fact set $\mathcal F_{\pm}$ and the relation set $\mathcal R_{\pm}=\mathcal R\cup\mathcal R^{-1}$, where $\mathcal R^{-1}$ is the set of inverse relations.We further add a self-loop $(v,r_{\mathrm{self}},v)$ to each entity, yielding the fact set $\widetilde{\mathcal F}$ and the relation set $\overline{\mathcal R}=\mathcal R_{\pm}\cup{r_{\mathrm{self}}}$ for entity-level message passing. Rules are mined exclusively from $\mathcal F_{\pm}$, excluding the artificially added entity self-loops. QORH first generates query-conditioned base relation representations $\mathbf z_{r\mid q}\in\mathbb R^d$ on the relation dependency graph, then obtains enhanced representations $\widetilde{\mathbf z}_{r\mid q}$ through the rule hypergraph, and finally uses these enhanced representations to perform entity-level message passing from the head entity $h$ and score candidate tail entities. Here, $d$ is the hidden dimension, and the subscript $r\mid q$ indicates that the representation of relation $r$ is conditioned on the query relation $q$.

\subsection{Query-Conditioned Relation Encoding}
As shown in Figure~\ref{fig:methodology_overview}(a), QORH first extracts local relational context from the observed graph to obtain base representations for the subsequent encoding of complete rules. The base relation encoder follows the relation graph modeling approaches used in prior work~\cite{leeINGRAMInductiveKnowledge2023,galkinFoundationModelsKnowledge2024,duGraphOracleEfficientFullyInductive2026a}, treating relation types as nodes and converting local connectivity patterns in the knowledge graph into structural links between relations. Specifically, we construct a directed relation graph based on whether two entity-level edges form a head-to-tail path through an intermediate entity:
\begin{equation}
    \mathcal{E}_R^{\rightarrow}=\left\{(r,s)\in\mathcal{R}_{\pm}^{2}\;\middle|\;\exists u,x,v\in\mathcal{V}:(u,r,x)\in\mathcal{F}_{\pm}\land(x,s,v)\in\mathcal{F}_{\pm}\right\}.
\end{equation}

Here, $r$ and $s$ are the relations of the first and second edges along the path, respectively, and $x$ is the intermediate entity connecting the two edges. To enable information exchange between adjacent relations, we augment this dependency set with reverse propagation edges and add a self-loop to each relation node, yielding the relation graph $\mathcal{G}_R=(\overline{\mathcal{R}},\mathcal{E}_R)$, where
\begin{equation}
    \mathcal{E}_R=\mathcal{E}_R^{\rightarrow}\cup\left\{(s,r)\mid(r,s)\in\mathcal{E}_R^{\rightarrow}\right\}\cup\left\{(r,r)\mid r\in\overline{\mathcal{R}}\right\}.
\end{equation}

These reverse propagation edges connect relation nodes and are distinct from the inverse relations that represent inverse facts in the entity graph. Inspired by the query-conditioned relation dependency graph modeling approach in GraphOracl~\cite{duGraphOracleEfficientFullyInductive2026a}e, QORH initializes relation nodes with query indicators for a given query relation $q$:
\begin{equation}
    \mathbf{z}_{r\mid q}^{(0)}=
    \begin{cases}\mathbf 1_d,&r=q,\\
    \mathbf 0_d,&r\neq q,
    \end{cases}
    \qquad r\in\overline{\mathcal R},
\end{equation}
where $\mathbf 1_d$ and $\mathbf 0_d$ are the $d$-dimensional all-ones and zero vectors, respectively. This initialization provides an initial signal at the query relation for structural propagation, allowing different query relations to obtain their own contextual representations. Let $\mathcal N_R(s)=\{r\mid(r,s)\in\mathcal E_R\}$ denote the set of incoming neighbors of relation node $s$, including $s$ itself. At propagation layer $k$, the attention score and normalized weight for edge $(r,s)$ are given by
\begin{equation}
\begin{aligned}
    e_{r,s\mid q}^{(k)} &= \left(\mathbf a_R^{(k)}\right)^\top \left[\mathbf{W}_A^{(k)}\mathbf{z}_{r\mid q}^{(k-1)} \;\Vert\; \mathbf{W}_A^{(k)}\mathbf{z}_{s\mid q}^{(k-1)}\right],\\
    \alpha_{r,s\mid q}^{(k)} &= \frac{\exp\!\left(e_{r,s\mid q}^{(k)}\right)} {\displaystyle\sum_{u\in\mathcal{N}_R(s)}\exp\!\left(e_{u,s\mid q}^{(k)}\right)}.
\end{aligned}
\end{equation}

Here, $\Vert$ denotes vector concatenation, $\mathbf{W}_A^{(k)}\in\mathbb{R}^{d\times d}$ is the attention projection matrix, and $\mathbf{a}_R^{(k)}\in\mathbb{R}^{2d}$ is a learnable attention vector. Normalization is performed over the incoming edges of the receiving relation $s$ for a fixed query $q$. The relation representation is updated as
\begin{equation}
    \mathbf{z}_{s\mid q}^{(k)} = \operatorname{ReLU}\!\left(\mathbf W_O^{(k)}\left[\mathbf W_N^{(k)} \sum_{r\in\mathcal N_R(s)} \alpha_{r,s\mid q}^{(k)} \mathbf z_{r\mid q}^{(k-1)} + \mathbf W_S^{(k)} \alpha_{s,s\mid q}^{(k)} \mathbf{z}_{s\mid q}^{(k-1)} \right] \right).
\end{equation}

Here, $\mathbf{W}_N^{(k)}$, $\mathbf{W}_S^{(k)}$, and $\mathbf{W}_O^{(k)}$ are learnable $d\times d$ matrices for neighborhood aggregation, transformation of the node's own information, and output mapping, respectively; $\operatorname{ReLU}$ is an element-wise activation function. Neighborhood aggregation includes self-loop messages, while a separate transformation of the node's own information preserves its query-dependent state. After $L_R$ propagation layers, we set $\mathbf{z}_{r\mid q}=\mathbf{z}_{r\mid q}^{(L_R)}$, where $L_R$ is the depth of the relation encoder. The resulting representation captures the local structural context of relation $r$ relative to query $q$ and serves as the input to the subsequent rule hypergraph encoder.

\subsection{Query-Specific Rule Hypergraph Construction}
The relation graph provides QORH with query-dependent local context, but its pairwise edges do not explicitly indicate whether multiple dependencies originate from the same complete path. To address this limitation, QORH further mines ordered path rules from the observed graph and represents each selected rule in its entirety as a hyperedge. Figure~\ref{fig:methodology_overview}(b) illustrates the process from rule mining to learnable selection.

\subsubsection{Mining Ordered Path Rules}
To associate complete paths with their conclusion relations, QORH mines ordered path rules of the following form from the observed facts:
\begin{equation}
    \rho:\quad r_1(x_0,x_1) \land r_2(x_1,x_2) \land \cdots \land r_{\ell_\rho}(x_{\ell_\rho-1},x_{\ell_\rho}) \Rightarrow q(x_0,x_{\ell_\rho}).
\end{equation}

Here, $\rho$ denotes a rule, $q$ is its head relation, $(r_1,\ldots,r_{\ell_\rho})$ is its ordered body, and $\ell_\rho$ is its body length, satisfying $1 \leq \ell_\rho \leq L_{\max}$; $L_{\max}$ is a configurable maximum rule length. The variables $x_0,\ldots,x_{\ell_\rho}$ denote entities along the path, and adjacent relations in the rule body are linked through shared variables. Thus, a rule specifies not only which relations co-occur but also their composition order and the correspondence between their endpoints.

Specifically, we enumerate simple directed paths over $\mathcal{F}_{\pm}$ that do not revisit any entity and group paths with identical relation sequences into the same rule body pattern. A relation type may occur at multiple positions along a path. To control computational cost, we cap the number of outgoing edges retained per entity and the number of paths enumerated per starting entity at each depth, with default limits of 64 and 1024, respectively. Let $\mathcal P$ denote the set of enumerated paths; for a rule $\rho$, we define the set of endpoint entity pairs associated with its body and the set of head facts as

\begin{equation}
\begin{aligned}
    \mathcal B_\rho &= \left\{(v_0,v_{\ell_\rho}) \;\middle|\; (v_0,r_1,v_1,\ldots,r_{\ell_\rho},v_{\ell_\rho})\in\mathcal P \right\},\\
    \mathcal H_q &= \left\{(u,v)\mid(u,q,v)\in\mathcal F_{\pm}\right\}.
\end{aligned}
\end{equation}

Here, $v_0,\ldots,v_{\ell_\rho}$ are the entities in a concrete path; $\mathcal B_\rho$ contains the distinct entity pairs connected by the enumerated rule body paths, and $\mathcal H_q$ contains the distinct entity pairs directly connected by relation $q$. Computing statistics over sets of entity pairs ensures that each endpoint pair is counted only once, even when multiple paths instantiate the same rule body between the two endpoints. The support, confidence, and coverage of a rule are defined as
\begin{equation}
    s_\rho=|\mathcal B_\rho\cap\mathcal H_q|, \qquad \operatorname{conf}_\rho=\frac{s_\rho}{|\mathcal B_\rho|}, \qquad \operatorname{cov}_\rho=\frac{s_\rho}{|\mathcal H_q|}.
\end{equation}

Here, $s_\rho$ counts the endpoint entity pairs that satisfy both the rule body and the rule head, quantifying the observed support for the rule; $\operatorname{conf}_\rho$ measures the proportion of entity pairs connected by the rule body that also have the corresponding head fact; $\operatorname{cov}_\rho$ measures the proportion of observed head facts that can be explained by the rule body. These statistics characterize the amount of supporting evidence, conditional reliability, and coverage of observed facts, respectively. We retain only candidate rules whose support is at least the threshold $s_{\min}$ and exclude tautological identity rules of the form $q(x,y)\Rightarrow q(x,y)$. Let $\mathcal C_q$ denote the resulting set of candidate rules with head relation $q$. 

For example, support for $r_1(x,z)\land r_2(z,y)\Rightarrow q(x,y)$ requires observing both the complete two-hop path and relation $q$ for the same endpoint pair $(x,y)$. This unit of statistical analysis jointly retains the complete path and its conclusion relation, enabling subsequent modeling to exploit joint information that is absent from independent relation pairs.

\subsubsection{Learnable Rule Selection}
The mining procedure yields candidate rules with observed support, but these rules still vary in the amount of support, coverage, and path length.To align rule use with the link prediction objective, QORH introduces a learnable selector before hyperedge encoding. Specifically, we construct a four-dimensional feature vector for each candidate rule:
\begin{equation}
    \boldsymbol\phi_\rho = \left[\frac{s_{\rho}+1}{|\mathcal{B}_{\rho}|+2}, \frac{s_{\rho}+1}{|\mathcal{H}_{q}|+2}, \frac{\ell_{\rho}}{L_{\max}}, \frac{\log(1+s_{\rho})}{\log(1+s_{\max})} \right]^\top \in\mathbb{R}^4.
\end{equation}

Here, $s_{\max}$ is the maximum support among all candidate rules in the current observed graph, taken over the union of all nonempty candidate sets $\mathcal C_q$. The first two components are the confidence and coverage with add-one smoothing, respectively, which mitigates extreme estimates arising from limited observations; the last two components normalize rule length and support, respectively, with the logarithmic transformation reducing numerical differences across support scales.

The rule selection score is computed by a shared multilayer perceptron (MLP):
\begin{equation}
    \omega_\rho = \sigma\!\left(\mathbf{w}_{s2}^{\top}\operatorname{SiLU}\left(\mathbf W_{s1}\boldsymbol\phi_\rho+\mathbf{b}_{s1}\right)+b_{s2}\right).
\end{equation}

Here, $\mathbf W_{s1}\in\mathbb R^{d_s\times4}$, $\mathbf b_{s1}\in\mathbb R^{d_s}$, $\mathbf w_{s2}\in\mathbb R^{d_s}$, and $b_{s2}\in\mathbb R$ are learnable parameters, and $d_s$ is the hidden dimension of the selector; $\sigma$ is the sigmoid function, and $\operatorname{SiLU}(x)=x\sigma(x)$ is the hidden-layer activation function; $\omega_\rho\in(0,1)$ denotes the rule selection score.

For query relation $q$, we select up to $K$ rules with the highest scores from $\mathcal C_q$:
\begin{equation}
    k_q=\min(K,|\mathcal{C}_q|), \qquad \mathcal{S}_q = \operatorname{TopK}_{k_q}\left(\mathcal{C}_q;\omega_\rho\right).
\end{equation}

Here, $K$ is a configurable upper bound on the number of selected rules, $k_q$ is the actual number selected, and $\mathcal S_q$ is the selected rule set; $\operatorname{TopK}$ selects rules in descending order of $\omega_\rho$. If $\mathcal C_q$ is empty, we set $\mathcal S_q=\varnothing$.

Rule selection precedes hyperedge encoding, restricting the more computationally expensive sequence encoding and hypergraph propagation to the selected rules. The selection score $\omega_\rho$ also contributes to hyperedge feature encoding and message attention, linking rule filtering and the use of rule information through the same score.

\subsubsection{Order- and Role-Preserving Hyperedge Representation}
After selecting the rules, QORH converts these discrete rules into hyperedge representations suitable for neural message passing. For each $\rho\in\mathcal{S}_q$, we construct a hyperedge connecting the rule head and body relations. Recording only the relations contained in a hyperedge is insufficient to distinguish different ordered rules, so we also preserve each relation's position and role within the rule.

We define
\begin{equation}
    r(\rho,j) = 
    \begin{cases}
    q,&j=0,\\
    r_j,&1\leq j\leq\ell_\rho,
    \end{cases}
    \quad
    \mathcal{I}_q =\left\{\bigl(r(\rho,j),\rho,j\bigr)\;\middle|\;\rho\in\mathcal{S}_q,\ 0 \leq j \leq \ell_\rho \right\}.
\end{equation}

Here, $j=0$ denotes the rule head role, while $j\geq1$ denotes position $j$ in the rule body; $r(\rho,j)$ returns the relation node associated with that role; $\mathcal I_q$ is the set of role-labeled relation-hyperedge incidences.

The rule hypergraph for the query relation is denoted by $\mathcal{G}_{H,q}=(\overline{\mathcal{R}},\mathcal{S}_q,\mathcal{I}_q)$, with relation types as nodes and a one-to-one correspondence between hyperedges and selected rules. Even when the same relation occurs at multiple positions in a rule body, the incidence set retains each position separately, preserving the distinctions between roles within the rule. Figure~\ref{fig:methodology_overview}(c) illustrates the construction of hyperedge representations, in which the rule body branch first encodes the ordered composition of relations. For the relation at body position $j$ of rule $\rho$, we construct a position-dependent input and sequentially update the hidden state using a gated recurrent unit (GRU):
\begin{equation}
    \begin{aligned}
    \mathbf{x}_{\rho,j} &= \mathbf{W}_B \mathbf{z}_{r_j\mid q}+\mathbf{p}_j,\\
    \mathbf b_{\rho,j} &= \operatorname{GRU}_B\left(\mathbf{x}_{\rho,j},\mathbf{b}_{\rho,j-1}\right), \qquad \mathbf{b}_{\rho,0}=\mathbf{0}_d,\\
    \mathbf{b}_\rho &= \mathbf{b}_{\rho,\ell_\rho}.
    \end{aligned}
\end{equation}

Here, $\mathbf{W}_B\in\mathbb{R}^{d\times d}$ is a projection matrix, $\mathbf{p}_j\in\mathbb{R}^d$ is the learnable embedding for body position $j$, and $\mathbf{x}_{\rho,j}\in\mathbb{R}^d$ is the input feature at that position; $\operatorname{GRU}_B$ is a sequence encoder shared across all rules, $\mathbf{b}_{\rho,j}$ is its hidden state after processing the first $j$ relations, and $\mathbf{b}_\rho$ is the representation of the complete rule body. Encoding is applied only to valid positions within each rule. Positional embeddings identify each relation's position along the path, while recurrent updates compose relation information in path order.

We then encode the rule statistics and selection score into a statistical representation and fuse it with the head and body representations:
\begin{equation}
\begin{aligned}
    \mathbf{c}_{\rho} &= \operatorname{MLP}_{\mathrm{stat}}\left([\boldsymbol\phi_\rho\Vert\omega_\rho]\right),\\
    \mathbf{e}_{\rho} &= \operatorname{LN}_{e}\left(\operatorname{MLP}_{\mathrm{edge}} \left([\mathbf z_{q\mid q}\Vert\mathbf b_\rho\Vert\mathbf c_\rho] \right) \right). 
\end{aligned}
\end{equation}

Here, $\operatorname{MLP}_{\mathrm{stat}}:\mathbb R^5\rightarrow\mathbb R^d$ maps the four statistical features and the selection score to $\mathbf c_\rho\in\mathbb R^d$; $\operatorname{MLP}_{\mathrm{edge}}:\mathbb R^{3d}\rightarrow\mathbb R^d$ fuses the three types of input, with both MLPs implemented as two-layer perceptrons with GELU activation and hidden dimension $d$; $\operatorname{LN}_e$ denotes layer normalization; $\mathbf e_\rho\in\mathbb R^d$ is the resulting hyperedge representation. The three components provide the query relation targeted by the rule, the ordered composition pattern of its body, and the statistical evidence for the rule, respectively. QORH thus encodes each rule into a unified representation incorporating its head, body, and statistical evidence. For example, the representation of rule $r_1(x,z)\land r_2(z,y)\Rightarrow q(x,y)$ preserves the role of $q$ as the conclusion, the relative positions of $r_1$ and $r_2$, and the statistics of their joint support.

The hyperedge representation specifies how a complete rule is encoded; the next step is to determine how this information contributes to relation nodes serving different roles.

\subsection{Role-Aware Hypergraph Message Passing}
To propagate complete rule information back to the relation level, QORH performs hyperedge-to-relation message passing along role-labeled incidences, which, together with the relation-to-hyperedge encoding in the preceding section, constitutes the relation enhancement process. As shown in Figure~\ref{fig:methodology_overview}(d), this process first generates messages according to rule roles, then aggregates them using incidence compatibility and rule selection scores, and finally updates relation representations through gated residual connections.

\subsubsection{Role-Specific Messages and Rule-Quality Attention}
The head and body relations of a rule serve different functions in the reasoning structure, and different positions within the body correspond to different stages of relation composition. QORH therefore uses role-specific transformations to map the same hyperedge representation to messages for different participating relations. For a relation-hyperedge incidence $(r(\rho,j),\rho,j)$, we define
\begin{equation}
    \boldsymbol\mu_{\rho,j} = \mathbf W_j^{\mathrm{msg}}\mathbf{e}_\rho, \qquad \mathbf W_j^{\mathrm{msg}} = 
    \begin{cases}
        \mathbf{W}_{\mathrm{head}},&j=0,\\
        \mathbf W_{\mathrm{body},j},&1\leq j\leq L_{\max}.
    \end{cases}
\end{equation}

Here, $\boldsymbol\mu_{\rho,j}\in\mathbb R^d$ is the message sent by rule $\rho$ to the relation in role $j$; $\mathbf{W}_{\mathrm{head}}\in\mathbb{R}^{d\times d}$ is the message transformation for the rule head, and $\mathbf{W}_{\mathrm{body},j}\in\mathbb{R}^{d\times d}$ is the message transformation for body position $j$. A relation may participate in multiple rules, so its incoming messages should account for rule content, the state of the receiving relation, and rule quality. To this end, we combine relation-hyperedge incidence compatibility with the selection score to compute attention:
\begin{equation}
    \xi_{\rho,j\mid q} = \mathbf{w}_H^\top\tanh\!\left(\mathbf{W}_{Hr}\mathbf{z}_{r(\rho,j)\mid q} + \mathbf{W}_{He}\mathbf{e}_\rho + \mathbf{u}_j\right) + \eta\log\!\left(\max(\omega_\rho,\varepsilon)\right).
\end{equation}

Here, $\xi_{\rho,j\mid q}$ denotes the unnormalized attention score, conditioned on query relation $q$, for the relation–hyperedge incidence indexed by $(\rho,j)$. The matrices $\mathbf W_{Hr},\mathbf W_{He}\in\mathbb R^{d\times d}$ map the representation of the receiving relation node and the rule hyperedge representation, respectively, into a shared attention feature space. The symbol $\rho$ denotes the rule associated with the incidence, and $j$ denotes the role index, with $j=0$ corresponding to the rule head and $j\geq1$ to the $j$-th position in the rule body. The notation $r(\rho,j)$ identifies the relation node associated with that role, and $\mathbf z_{r(\rho,j)\mid q}\in\mathbb R^d$ denotes its base representation conditioned on query relation $q$, before hypergraph enhancement. The vector $\mathbf e_\rho\in\mathbb R^d$ represents the hyperedge corresponding to rule $\rho$, while $\mathbf u_j\in\mathbb R^d$ is a role embedding used to distinguish a relation’s position and function within the rule when computing compatibility. The vector $\mathbf w_H\in\mathbb R^d$ is the attention scoring vector, and $\tanh$ denotes the element-wise hyperbolic tangent function. The rule-selection score $\omega_\rho\in(0,1)$ is produced by the rule selector, $\eta\geq0$ controls its strength as an attention prior, and $\varepsilon=10^{-8}$ ensures numerical stability in the logarithm computation. This logarithmic prior allows rule-selection scores to further modulate the relative contributions of incidence messages beyond content compatibility.

For a given query relation $q$ and receiving relation node $r$, let $\mathcal I_q(r)=\{(\rho,j)\mid(r,\rho,j)\in\mathcal I_q\}$ denote the set of rule-role pairs incident to that node. If $\mathcal I_q(r)$ is nonempty, we apply softmax normalization to the attention scores of these incidences and aggregate the corresponding role-specific messages using the resulting weights:
\begin{equation}
    \begin{aligned}
        \beta_{\rho,j\mid q} &= \frac{\exp(\xi_{\rho,j\mid q})}{\displaystyle\sum_{(\rho',j')\in\mathcal I_q(r)} \exp(\xi_{\rho',j'\mid q})}, \qquad (\rho,j)\in\mathcal I_q(r),\\
        \mathbf m_{r\mid q} &= \sum_{(\rho,j)\in\mathcal I_q(r)} \beta_{\rho,j\mid q}\boldsymbol\mu_{\rho,j}.
    \end{aligned}
\end{equation}

Here, $\beta_{\rho,j\mid q}$ is the normalized attention weight for incidence $(\rho,j)$, satisfying $\sum_{(\rho,j)\in\mathcal I_q(r)}\beta_{\rho,j\mid q}=1$ at each receiving relation node; $\rho'$ and $j'$ index the rules and their roles, respectively, in the normalization sum; $\boldsymbol\mu_{\rho,j}\in\mathbb R^d$ is the message generated by rule $\rho$ for role $j$; $\mathbf m_{r\mid q}\in\mathbb R^d$ is the aggregated rule message for relation node $r$ conditioned on query relation $q$. Normalization is performed over all rule-role pairs incident to the same receiving relation node under the same query relation. When a relation occurs at multiple positions within a rule, each occurrence is treated as a separate incidence during attention normalization and message aggregation, preserving the contributions of different roles to the relation update. Both head and body relations receive rule information through their respective role-labeled incidences.

\subsubsection{Gated Relation Update}
The aggregated message provides complementary information from complete rules for the current relation, while the base relation representation retains its local structural context. To combine these two sources of information, we introduce an element-wise update gate:
\begin{equation}
    \begin{aligned}
    \mathbf g_{r\mid q} &= \sigma\!\left(\mathbf W_g[\mathbf z_{r\mid q}\Vert\mathbf m_{r\mid q}]+\mathbf b_g\right),  \qquad \mathcal I_q(r)\neq\varnothing,\\
    \widetilde{\mathbf z}_{r\mid q} &=
    \begin{cases}
        \operatorname{LN}_u\left(\mathbf z_{r\mid q} + \mathbf g_{r\mid q}\odot\mathbf m_{r\mid q}\right),&\mathcal I_q(r)\neq\varnothing,\\
        \mathbf z_{r\mid q}, &\mathcal I_q(r)=\varnothing.
    \end{cases}
    \end{aligned}
\end{equation}

Here, $\mathbf W_g\in\mathbb R^{d\times2d}$ and $\mathbf b_g\in\mathbb R^d$ are the gating parameters, $\mathbf g_{r\mid q}\in(0,1)^d$ controls the extent to which rule messages are incorporated into each feature dimension, $\odot$ denotes element-wise multiplication, $\operatorname{LN}_u$ is the layer normalization used in the relation update, and $\widetilde{\mathbf z}_{r\mid q}$ is the final enhanced relation representation. The residual term preserves the context provided by the base relation graph, while the gated term introduces complete rule information relevant to the current relation. Relations that do not participate in any selected rule retain their original representations; when $\mathcal S_q=\varnothing$, all relations retain the outputs of the base relation encoder. The resulting representations $\widetilde{\mathbf z}_{r\mid q}$ serve as both edge relation and query relation representations in entity reasoning, allowing relation-level rule information to influence message passing between individual entities.

\subsection{Rule-Enhanced Entity-Level Reasoning}
As shown in Figure~\ref{fig:methodology_overview}(d), given a query $(h,q,?)$, QORH uses the enhanced relation representations to perform $L_E$ layers of message passing over $\widetilde{\mathcal F}$, starting from the head entity $h$. Let $\mathcal A^{(\ell)}$ denote the set of active entities at layer $\ell$, defined as
\begin{equation}
    \mathcal A^{(0)}=\{h\}, \qquad \mathcal A^{(\ell)} = \left\{v\in\mathcal V \;\middle|\; \exists u\in\mathcal A^{(\ell-1)},\ r\in\overline{\mathcal R}:(u,r,v)\in\widetilde{\mathcal F}\right\}.
\end{equation}

Here, $\ell=1,\ldots,L_E$, and entity self-loops ensure that previously active nodes continue to participate in message passing. Let $\mathbf h_v^{(\ell)}\in\mathbb R^d$ denote the hidden state of entity $v$ at layer $\ell$; we omit its conditioning subscript $(h,q)$ for brevity and set $\mathbf h_h^{(0)}=\mathbf 0_d$. For each edge $(u,r,v)$ with $u\in\mathcal A^{(\ell-1)}$, we compute its message and query-conditioned propagation weight from the source entity state and the enhanced relation representations:
\begin{equation}
    \begin{aligned}
        \mathbf m_{u,r,v}^{(\ell)} &= \mathbf h_u^{(\ell-1)} + \widetilde{\mathbf z}_{r\mid q},\\
        \mathbf a_{u,r\mid q}^{(\ell)} &= \operatorname{ReLU}\!\left( \mathbf W_s^{(\ell)}\mathbf h_u^{(\ell-1)} + \mathbf W_r^{(\ell)}\widetilde{\mathbf z}_{r\mid q} + \mathbf W_q^{(\ell)}\widetilde{\mathbf z}_{q\mid q} + \mathbf b_a^{(\ell)} \right),\\
        \gamma_{u,r,v}^{(\ell)} &= \sigma\!\left( \left(\mathbf w_\alpha^{(\ell)}\right)^\top \mathbf a_{u,r\mid q}^{(\ell)} + b_\alpha^{(\ell)} \right).
    \end{aligned}
\end{equation}

Here, $\mathbf m_{u,r,v}^{(\ell)}\in\mathbb R^d$ is the message on an entity-level edge, $\mathbf a_{u,r\mid q}^{(\ell)}\in\mathbb R^{d_a}$ is the query-conditioned attention feature, and $\gamma_{u,r,v}^{(\ell)}\in(0,1)$ is the propagation weight of that edge; the matrices $\mathbf W_s^{(\ell)},\mathbf W_r^{(\ell)},\mathbf W_q^{(\ell)}\in\mathbb R^{d_a\times d}$ map the source entity state, edge relation representation, and query relation representation, respectively, $\mathbf b_a^{(\ell)},\mathbf w_\alpha^{(\ell)}\in\mathbb R^{d_a}$ and $b_\alpha^{(\ell)}\in\mathbb R$ are bias and output parameters, and $d_a$ is the hidden dimension of attention over entity-level edges; the edge weights $\gamma_{u,r,v}^{(\ell)}\in(0,1)$ are computed independently using the sigmoid function, allowing multiple query-relevant edges to transmit information simultaneously.

The target entity then aggregates messages from nodes active at the preceding layer and updates its state using a GRU operating across layers:
\begin{equation}
    \overline{\mathbf h}_v^{(\ell)} = \varphi\!\left(\mathbf W_h^{(\ell)} \sum_{\substack{u\in\mathcal A^{(\ell-1)}\\ (u,r,v)\in\widetilde{\mathcal F} }} \gamma_{u,r,v}^{(\ell)}\mathbf m_{u,r,v}^{(\ell)}\right),
\end{equation}

\begin{equation}
    \mathbf h_v^{(\ell)} = \operatorname{GRU}_E \left(\operatorname{Dropout} \left(\overline{\mathbf h}_v^{(\ell)}\right), \widehat{\mathbf h}_v^{(\ell-1)}\right).
\end{equation}

Here, $\mathbf W_h^{(\ell)}\in\mathbb R^{d\times d}$ is a learnable transformation, and $\varphi$ is an element-wise activation function. For previously active entities, $\widehat{\mathbf h}_v^{(\ell-1)}$ is their hidden state from the preceding layer, whereas it is the zero vector for entities newly activated at the current layer; $\operatorname{GRU}_E$ shares parameters across entity reasoning layers and is parameterized independently of the rule body encoder $\operatorname{GRU}_B$, while $\operatorname{Dropout}$ provides regularization during training.

After $L_E$ propagation layers, a shared learnable scoring vector $\mathbf w_o\in\mathbb R^d$ is used to compute candidate tail entity scores:
\begin{equation}
    s_\theta(h,q,t\mid\mathcal G_{\mathrm{obs}}) = 
    \begin{cases}
        \mathbf w_o^\top\mathbf h_t^{(L_E)}, &t\in\mathcal A^{(L_E)},\\
        0, &t\notin\mathcal A^{(L_E)}.
    \end{cases}
\end{equation}

\subsection{Optimization and Inductive Inference}
The entity-inductive and fully inductive experiments for QORH follow the data splits and evaluation protocols of GraIL~\cite{teruInductiveRelationPrediction2020} and InGram~\cite{leeINGRAMInductiveKnowledge2023}, respectively. During training, we partition the training facts into message-passing facts and supervision queries and update the relation dependency graph and rule base as the observed facts change. Model parameters are optimized exclusively on training data, with the validation set used for model selection and the test set reserved for final evaluation. Given a batch of supervision triples $\mathcal D_b$, we apply softmax normalization to the scores over the set of training candidate entities $\mathcal V_{\mathrm{tr}}$ and minimize the negative log-likelihood loss:
\begin{equation}
    \begin{aligned} 
        p_\theta(t\mid h,q,\mathcal G_{\mathrm{obs}}) &= \frac{\exp\!\left(s_\theta(h,q,t\mid\mathcal G_{\mathrm{obs}})\right)}{\displaystyle\sum_{v\in\mathcal V_{\mathrm{tr}}} \exp\!\left(s_\theta(h,q,v\mid\mathcal G_{\mathrm{obs}})\right)},\\
        \mathcal L(\theta) &= -\sum_{(h,q,t)\in\mathcal D_b} \log p_\theta(t\mid h,q,\mathcal G_{\mathrm{obs}}). 
    \end{aligned}
\end{equation}

Here, $p_\theta$ is the conditional predictive probability of a candidate tail entity, $t$ is the correct tail entity, $v$ ranges over all training candidate entities, and $\mathcal L$ is the batch loss. All candidate entities are included in the normalization, eliminating the need for additional negative sampling. During inductive inference, with the parameters $\theta$ fixed, we reconstruct the relation graph, mine rules, and compute their statistics from the observed facts in the inference graph before scoring the candidate entities in $\mathcal V_{\mathrm{inf}}$. Because the shared parameters are independent of specific entity or relation IDs, the model can generate representations for unseen entities and relations from the structure of a new graph.

To avoid redundant computation, queries with the same query relation within a batch share the relation encoding and hypergraph enhancement results; candidate rules and their statistics can also be reused when the observed graph remains unchanged. For each query relation, at most $K$ rules with body length no greater than $L_{\max}$ are selected, and the total number of incidences satisfies $|\mathcal I_q|=\sum_{\rho\in\mathcal S_q}(\ell_\rho+1)\leq K(L_{\max}+1)$, bounding the scale of hypergraph propagation through the limits on rule count and length.

\section{Experiments}
This section evaluates QORH on entity-inductive and fully inductive knowledge graph link prediction tasks, covering 12 datasets in each setting. Using mean reciprocal rank (MRR) as the primary evaluation metric, QORH achieves the best performance among the compared methods on 11 of the 12 entity-inductive datasets and all 12 fully inductive datasets, demonstrating its strong link prediction performance in both inductive settings.

\subsection{Datasets and Evaluation Metrics}

\subsubsection{Entity-Inductive Datasets}
We adopt the entity-inductive link prediction benchmarks introduced by GraIL~\cite{teruInductiveRelationPrediction2020}. These benchmarks comprise three dataset families derived from WN18RR~\cite{dettmersConvolutional2DKnowledge2018}, FB15k-237~\cite{toutanovaObservedLatentFeatures2015}, and NELL-995~\cite{xiongDeepPathReinforcementLearning2017}, each containing four versions, v1 through v4, yielding 12 datasets in total. Each version consists of a training graph and a corresponding inductive inference graph, enabling evaluation of generalization across different graph sizes and relational structures. Under the benchmark protocol, the training and inference graphs must have disjoint entity sets, while inference uses the relation vocabulary available during training, i.e., $\mathcal V_{\mathrm{tr}}\cap\mathcal V_{\mathrm{inf}}=\varnothing, \quad \mathcal R_{\mathrm{inf}}\subseteq\mathcal R_{\mathrm{tr}}.$

\subsubsection{Fully Inductive Datasets}
We further adopt the benchmarks introduced by InGram~\cite{leeINGRAMInductiveKnowledge2023}, which comprise three dataset families, FB, NL, and WK, derived from FB15k-237~\cite{toutanovaObservedLatentFeatures2015}, NELL-995~\cite{xiongDeepPathReinforcementLearning2017}, and Wikidata68K~\cite{geseseRAILDLeveragingRelation2022}, respectively. Each family includes four settings, 25, 50, 75, and 100, yielding 12 datasets in total. These numbers specify the nominal proportions of triples involving unseen relations during dataset construction, rather than the proportions of unseen relation types; the actual proportions in the data files may deviate from these nominal values. In the first three settings, the inference graphs contain both seen and unseen relations, whereas the 100 setting enforces completely disjoint relation sets between the training and inference graphs. Following the task definition in Section~\ref{sec:Problem Formulation and Framework Overview}, we collectively refer to these settings that allow unseen relations as fully inductive settings. Each dataset contains \texttt{train.txt}, \texttt{msg.txt}, \texttt{valid.txt}, and \texttt{test.txt}. The \texttt{train.txt} file is used for model training. The remaining three files belong to an inference graph whose entities are disjoint from those on the training side and provide observed facts, validation queries, and test queries, respectively.

\subsubsection{Evaluation Metrics}
\label{sec:Evaluation Metrics}

We evaluate link prediction performance using mean reciprocal rank (MRR), Hits@1, and Hits@10. For each test triple, we perform both head and tail entity prediction. Head prediction is converted into tail prediction using inverse relations, consistent with the inference formulation in Section~\ref{sec:Methodology}. Let $N$ denote the total number of prediction instances evaluated in both directions, and let $\operatorname{rank}_i$ denote the rank of the target entity for the $i$-th instance. The metrics are defined as follows:
\begin{equation}
    \operatorname{MRR} =\frac{1}{N}\sum_{i=1}^{N}\frac{1}{\operatorname{rank}_i},
\end{equation}

\begin{equation}
    \operatorname{Hits}@k =\frac{1}{N}\sum_{i=1}^{N} \mathbb{I}\!\left(\operatorname{rank}_i\leq k\right), \qquad k\in\{1,10\},
\end{equation}
where $\mathbb{I}(\cdot)$ is the indicator function. MRR measures the ranking positions of correct answers and is particularly sensitive to changes near the top of the candidate list. Hits@1 and Hits@10 measure the proportions of correct answers ranked first and within the top ten, respectively. Higher values indicate better performance for all three metrics.

To comprehensively evaluate ranking performance over the candidate entity space, we use filtered evaluation with all entities as candidates. For a query $(h,q,?)$, all entities in the current message-passing graph are considered as candidates and ranked according to $s_\theta(h,q,t\mid\mathcal G_{\mathrm{obs}})$. When computing the target entity’s rank, we filter out other candidate entities that form known true triples with the query, preventing valid alternative answers from being counted as incorrect predictions. Accordingly, no negative sampling is performed during evaluation. Compared with the original experimental settings of GraIL~\cite{teruInductiveRelationPrediction2020}, MorsE~\cite{chenMetaKnowledgeTransferInductive2022}, and related methods, which rank each test triple against 50 randomly sampled negative triples, all-entity ranking covers the complete candidate space and avoids the randomness introduced by candidate sampling. All MRR, Hits@1, and Hits@10 scores reported for QORH are computed under this all-entity evaluation setting.

\subsection{Baselines and Experimental Setup}
\label{sec:Baselines and Experimental Setup}

To systematically evaluate QORH on entity-inductive and fully inductive link prediction tasks, we select 20 representative baseline methods spanning five methodological categories: logical rule learning, entity-graph reasoning, structural representation learning, semantic feature utilization, and relational structure transfer.

\begin{enumerate}
    \renewcommand{\labelenumi}{(\arabic{enumi})}

    \item \textbf{Logical rule learning methods:} RuleN~\cite{meilickeFineGrainedEvaluationRule2018}, Neural LP~\cite{yangDifferentiableLearningLogical2017}, and DRUM~\cite{sadeghianDRUMEndToEndDifferentiable2019}. These methods learn logical composition patterns among relations and predict missing facts through rule instantiation.

    \item \textbf{Entity-graph structure and path propagation methods:} GraIL~\cite{teruInductiveRelationPrediction2020}, CoMPILE~\cite{maiCommunicativeMessagePassing2021}, SNRI~\cite{xuSubgraphNeighboringRelations2022}, INDIGO~\cite{liuINDIGOGNNBasedInductive2021}, NBFNet~\cite{zhuNeuralBellmanfordNetworks2021}, RED-GNN~\cite{zhangKnowledgeGraphReasoning2022}, and Context Pooling~\cite{suContextPoolingQueryspecific2025}. These methods extract structural evidence from entity neighborhoods, local subgraphs, or multi-hop paths to learn transferable link prediction patterns.

    \item \textbf{Representation methods based on structural encoding and meta-learning:} CompGCN~\cite{vashishthCompositionbasedMultiRelationalGraph2020}, NodePiece~\cite{galkinNodePieceCompositionalParameterEfficient2021}, and MorsE~\cite{chenMetaKnowledgeTransferInductive2022}. These methods learn representations for link prediction through compositional encoding of entities and relations, shared representation units, or transferable mechanisms for generating embeddings.

    \item \textbf{Methods using semantic and auxiliary features:} BLP~\cite{dazaInductiveEntityRepresentations2021}, QBLP~\cite{aliImprovingInductiveLink2021}, and RAILD~\cite{geseseRAILDLeveragingRelation2022}. These methods incorporate textual descriptions, entity features, or fact qualifiers to provide additional semantic evidence for inductive prediction.

    \item \textbf{Relational structure transfer methods:} RMPI~\cite{gengRelationalMessagePassing2023}, InGram~\cite{leeINGRAMInductiveKnowledge2023}, TARGI~\cite{dingGlobalTopologyRelationGraph2025}, and GraphOracle~\cite{duGraphOracleEfficientFullyInductive2026a}. These methods explicitly model structural connections among relations and use relation-level message passing to support generalization to new entities and relations.
\end{enumerate}

Experiments are conducted on a server running Ubuntu 20.04 with an NVIDIA Tesla V100 GPU equipped with 32 GB of memory. QORH is implemented in PyTorch and uses torch\_scatter for sparse message aggregation. We use the Adam optimizer and set the random seed to 42. The hidden representation dimension is selected from $\{16,32,48,64\}$, the number of entity-level message-passing layers from {3,5}, and the batch size from {10,20,24,100}. The initial learning rate, $L_2$ regularization coefficient, and dropout rate range over $1.6\times10^{-4},9.2\times10^{-3}$, $1.8\times10^{-5},4.0\times10^{-4}$, and $0.02,0.30$, respectively. For the ordered rule hypergraph module, the maximum rule length is selected from $\{2, 3\}$, the minimum rule support is set to 2, and the number of rules selected for each query relation is chosen from $\{1,2,3,4,5,6,7\}$. QORH is trained for a maximum of 200 epochs, with validation MRR used for model selection and early stopping. Training terminates when validation MRR fails to improve for 10 consecutive epochs, and the model with the highest validation MRR is used for final testing.

As discussed in Sections~\ref{sec:Fully-Inductive Knowledge Graph Link Prediction} and~\ref{sec:Baselines and Experimental Setup}, the pretraining data of existing knowledge graph foundation models overlap with inductive test data, which is inconsistent with the inductive evaluation protocol. As shown in Table~\ref{tab:pretraining-overlap}, the official pretraining pipeline of GraphOracle~\cite{duGraphOracleEfficientFullyInductive2026a} includes source knowledge graphs such as FB15k-237~\cite{toutanovaObservedLatentFeatures2015} and WN18RR~\cite{dettmersConvolutional2DKnowledge2018}, from which some inductive benchmarks are also constructed. Consequently, even under zero-shot evaluation on the target dataset, a substantial fraction of the test entities and relations have already been encountered by the pretrained model.

\begin{table}[t]
    \centering
    \small
    \setlength{\tabcolsep}{2.5pt}
    \renewcommand{\arraystretch}{1.08}

    \begin{threeparttable}
        \caption{Overlap between GraphOracle's pretraining data and selected inductive test sets (\%).}
        \label{tab:pretraining-overlap}

        \begin{tabular}{@{}llrr@{}}
            \toprule
            \textbf{Inductive setting} &
            \textbf{Dataset} &
            \textbf{Entity overlap} &
            \textbf{Relation overlap} \\
            \midrule

            \multirow{8}{*}{Entity-inductive}
                & FB15k-237-v1 & 100.00 & \textemdash \\
                & FB15k-237-v2 &  96.98 & \textemdash \\
                & FB15k-237-v3 &  98.57 & \textemdash \\
                & FB15k-237-v4 &  98.88 & \textemdash \\
                & WN18RR-v1    &  83.92 & \textemdash \\
                & WN18RR-v2    &  81.41 & \textemdash \\
                & WN18RR-v3    &  71.18 & \textemdash \\
                & WN18RR-v4    &  82.03 & \textemdash \\

            \midrule

            \multirow{4}{*}{Fully inductive}
                & FB-25  & 97.34 &  97.45 \\
                & FB-50  & 98.25 &  97.66 \\
                & FB-75  & 95.45 &  97.35 \\
                & FB-100 & 95.78 & 100.00 \\

            \bottomrule
        \end{tabular}

        \begin{tablenotes}[flushleft]
            \footnotesize
            \item[] \textit{Note:}
            Entity and relation overlap indicate the percentages of distinct entities and relations in the inductive test set that appear in GraphOracle's pretraining data. GraphOracle's released data loader constructs training facts from \texttt{train.txt} without reading \texttt{facts.txt}, so overlap may be below 100\%. ``\textemdash'' denotes unreported relation overlap in the entity-inductive setting, which uses the training relation vocabulary.
        \end{tablenotes}
    \end{threeparttable}
\end{table}

To eliminate the effects of this data overlap, we strictly follow the official GraphOracle~\cite{duGraphOracleEfficientFullyInductive2026a} implementation(https://github.com/EnjunDu/GraphOracle) and adapt its data loading, training, and testing procedures to the entity-inductive protocol introduced by GraIL~\cite{teruInductiveRelationPrediction2020} and the fully inductive protocol introduced by InGram~\cite{leeINGRAMInductiveKnowledge2023}. Specifically, consistent with the other baselines, GraphOracle~\cite{duGraphOracleEfficientFullyInductive2026a} is trained independently on the designated training graph of each benchmark, with model selection based on the corresponding validation set. During testing, model parameters are fixed, observed facts in the corresponding inference graph are used for message passing, and prediction metrics are computed according to the evaluation procedure in Section~\ref{sec:Evaluation Metrics}.

\subsection{Main Results}
\label{sec:Main Results}

\begin{table*}[tbp]
    \centering
    \footnotesize
    \setlength{\tabcolsep}{2.5pt}
    \renewcommand{\arraystretch}{1.08}
    \resizebox{\linewidth}{!}{%
    \begin{threeparttable}
        \caption{Entity-inductive link prediction results on WN18RR, FB15k-237, and NELL-995. Bold and underlined values indicate the best and second-best results, respectively.}
        \label{tab:entity-inductive-results}

        \begin{tabular}{@{}ll*{12}{c}@{}}
            \toprule
            \multirow{2}{*}{\textbf{Metric}} & \multirow{2}{*}{\textbf{Models}}
            & \multicolumn{4}{c}{\textbf{WN18RR}}
            & \multicolumn{4}{c}{\textbf{FB15k-237}}
            & \multicolumn{4}{c}{\textbf{NELL-995}} \\
            \cmidrule(lr){3-6}\cmidrule(lr){7-10}\cmidrule(lr){11-14}
            & & \textbf{V1} & \textbf{V2} & \textbf{V3} & \textbf{V4}
            & \textbf{V1} & \textbf{V2} & \textbf{V3} & \textbf{V4}
            & \textbf{V1} & \textbf{V2} & \textbf{V3} & \textbf{V4} \\
            \midrule
            \multirow{12}{*}{\textbf{MRR}} & RuleN\tnote{*} & 0.668 & 0.645 & 0.368 & 0.624 & 0.363 & 0.433 & 0.439 & 0.429 & 0.615 & 0.385 & 0.381 & 0.333 \\
             & Neural LP\tnote{*} & 0.649 & 0.635 & 0.361 & 0.628 & 0.325 & 0.389 & 0.400 & 0.396 & 0.610 & 0.361 & 0.367 & 0.261 \\
             & DRUM\tnote{*} & 0.666 & 0.646 & 0.380 & 0.627 & 0.333 & 0.395 & 0.402 & 0.410 & 0.628 & 0.365 & 0.375 & 0.273 \\
             & GraIL\tnote{*} & 0.627 & 0.625 & 0.323 & 0.553 & 0.279 & 0.276 & 0.251 & 0.227 & 0.481 & 0.297 & 0.322 & 0.262 \\
             & CoMPILE\tnote{*} & 0.577 & 0.578 & 0.308 & 0.548 & 0.287 & 0.276 & 0.262 & 0.213 & 0.330 & 0.248 & 0.319 & 0.229 \\
             & NBFNet\tnote{*} & 0.684 & 0.652 & 0.425 & 0.604 & 0.307 & 0.369 & 0.331 & 0.305 & 0.584 & 0.410 & 0.425 & 0.287 \\
             & MorsE\tnote{\#} & 0.429 & 0.389 & 0.229 & 0.324 & 0.228 & 0.298 & 0.265 & 0.232 & 0.334 & 0.111 & 0.099 & 0.046 \\
             & RED-GNN\tnote{*} & 0.701 & 0.690 & \underline{0.427} & \underline{0.651} & 0.369 & \underline{0.469} & 0.445 & \underline{0.442} & 0.637 & 0.419 & \underline{0.436} & 0.363 \\
             & InGram\tnote{\#} & 0.279 & 0.235 & 0.114 & 0.230 & 0.192 & 0.223 & 0.188 & 0.162 & \textbf{0.672} & 0.219 & 0.278 & 0.203 \\
             & Context Pooling\tnote{*} & \underline{0.708} & \underline{0.694} & 0.425 & 0.648 & \underline{0.383} & 0.463 & \underline{0.447} & 0.427 & 0.634 & \underline{0.430} & 0.426 & \underline{0.382} \\
             & GraphOracle\tnote{\textasciicircum} & 0.656 & 0.634 & 0.393 & 0.490 & 0.362 & 0.307 & 0.407 & 0.219 & 0.504 & 0.360 & 0.404 & 0.262 \\
             & QORH & \textbf{0.793} & \textbf{0.725} & \textbf{0.490} & \textbf{0.679} & \textbf{0.466} & \textbf{0.533} & \textbf{0.517} & \textbf{0.497} & \underline{0.649} & \textbf{0.611} & \textbf{0.567} & \textbf{0.535} \\
            \midrule
            \multirow{12}{*}{\textbf{Hits@1}} & RuleN\tnote{*} & 0.635 & 0.611 & 0.347 & 0.592 & 0.309 & 0.347 & 0.345 & 0.338 & 0.545 & 0.304 & 0.303 & 0.248 \\
             & Neural LP\tnote{*} & 0.592 & 0.575 & 0.304 & 0.583 & 0.243 & 0.286 & 0.309 & 0.289 & 0.500 & 0.249 & 0.267 & 0.137 \\
             & DRUM\tnote{*} & 0.613 & 0.595 & 0.330 & 0.586 & 0.247 & 0.284 & 0.308 & 0.309 & 0.500 & 0.271 & 0.262 & 0.163 \\
             & GraIL\tnote{*} & 0.554 & 0.542 & 0.278 & 0.443 & 0.205 & 0.202 & 0.165 & 0.143 & 0.425 & 0.199 & 0.224 & 0.153 \\
             & CoMPILE\tnote{*} & 0.473 & 0.485 & 0.258 & 0.473 & 0.208 & 0.178 & 0.166 & 0.134 & 0.105 & 0.156 & 0.226 & 0.159 \\
             & NBFNet\tnote{*} & 0.592 & 0.575 & 0.304 & 0.574 & 0.190 & 0.229 & 0.206 & 0.185 & 0.500 & 0.271 & 0.262 & 0.233 \\
             & MorsE\tnote{\#} & 0.343 & 0.320 & 0.179 & 0.244 & 0.117 & 0.168 & 0.146 & 0.128 & 0.235 & 0.046 & 0.040 & 0.014 \\
             & RED-GNN\tnote{*} & 0.653 & 0.633 & \underline{0.368} & \underline{0.606} & 0.302 & \underline{0.381} & 0.351 & \underline{0.340} & 0.525 & 0.319 & \underline{0.345} & 0.259 \\
             & InGram\tnote{\#} & 0.146 & 0.125 & 0.056 & 0.100 & 0.093 & 0.127 & 0.106 & 0.087 & \textbf{0.570} & 0.113 & 0.189 & 0.125 \\
             & Context Pooling\tnote{*} & \underline{0.660} & \underline{0.637} & 0.365 & 0.603 & \underline{0.316} & 0.368 & \underline{0.358} & 0.330 & 0.517 & \underline{0.329} & 0.324 & \underline{0.278} \\
             & GraphOracle\tnote{\textasciicircum} & 0.516 & 0.535 & 0.326 & 0.289 & 0.273 & 0.189 & 0.293 & 0.116 & 0.455 & 0.228 & 0.326 & 0.131 \\
             & QORH & \textbf{0.753} & \textbf{0.678} & \textbf{0.423} & \textbf{0.638} & \textbf{0.381} & \textbf{0.443} & \textbf{0.425} & \textbf{0.399} & \underline{0.550} & \textbf{0.507} & \textbf{0.466} & \textbf{0.415} \\
            \midrule
            \multirow{12}{*}{\textbf{Hits@10}} & RuleN\tnote{*} & 0.730 & 0.694 & 0.407 & 0.681 & 0.446 & 0.599 & 0.600 & 0.605 & 0.760 & 0.514 & 0.531 & 0.484 \\
             & Neural LP\tnote{*} & 0.772 & 0.749 & 0.476 & 0.706 & 0.468 & 0.586 & 0.571 & 0.593 & 0.871 & 0.564 & 0.576 & 0.539 \\
             & DRUM\tnote{*} & 0.777 & 0.747 & 0.477 & 0.702 & 0.474 & 0.595 & 0.571 & 0.593 & \underline{0.873} & 0.540 & 0.577 & 0.531 \\
             & GraIL\tnote{*} & 0.760 & 0.776 & 0.409 & 0.687 & 0.429 & 0.424 & 0.424 & 0.389 & 0.565 & 0.496 & 0.518 & 0.506 \\
             & CoMPILE\tnote{*} & 0.747 & 0.743 & 0.406 & 0.670 & 0.439 & 0.457 & 0.449 & 0.358 & 0.575 & 0.446 & 0.515 & 0.421 \\
             & NBFNet\tnote{*} & \underline{0.827} & \underline{0.799} & \underline{0.563} & 0.702 & 0.517 & 0.639 & 0.588 & 0.559 & 0.795 & 0.635 & 0.606 & \underline{0.591} \\
             & MorsE\tnote{\#} & 0.588 & 0.523 & 0.307 & 0.469 & 0.471 & 0.582 & 0.504 & 0.454 & 0.485 & 0.241 & 0.213 & 0.096 \\
             & RED-GNN\tnote{*} & 0.799 & 0.780 & 0.524 & 0.721 & 0.483 & 0.629 & 0.603 & \underline{0.621} & 0.866 & 0.601 & 0.594 & 0.556 \\
             & InGram\tnote{\#} & 0.524 & 0.449 & 0.225 & 0.456 & 0.373 & 0.426 & 0.340 & 0.311 & 0.845 & 0.423 & 0.464 & 0.365 \\
             & Context Pooling\tnote{\#} & 0.790 & 0.785 & 0.527 & \underline{0.723} & 0.473 & \underline{0.640} & 0.589 & 0.592 & 0.866 & \underline{0.636} & \underline{0.619} & 0.564 \\
             & GraphOracle\tnote{\textasciicircum} & 0.798 & 0.783 & 0.523 & 0.700 & \underline{0.517} & 0.536 & \underline{0.611} & 0.428 & 0.535 & 0.616 & 0.541 & 0.563 \\
             & QORH & \textbf{0.878} & \textbf{0.801} & \textbf{0.603} & \textbf{0.752} & \textbf{0.600} & \textbf{0.694} & \textbf{0.677} & \textbf{0.676} & \textbf{0.950} & \textbf{0.792} & \textbf{0.745} & \textbf{0.741} \\

            \bottomrule
        \end{tabular}

        \begin{tablenotes}[flushleft]
            \footnotesize
            \item[*] Results taken directly from \cite{suContextPoolingQueryspecific2025} or \cite{duGraphOracleEfficientFullyInductive2026a}.

            \item[\#] Results reproduced using the code released in the respective official repositories, following the training and testing protocols for the entity-inductive setting.

            \item[\textasciicircum] Results reproduced by adapting the code released in the respective official repositories to the entity-inductive training and testing protocols, with modifications to the data loading, training, and evaluation procedures. See the \emph{Baselines and Experiment Setup} section for details.
        \end{tablenotes}
    \end{threeparttable}
    }%
\end{table*}

\begin{table*}[t]
    \centering
    \small
    \setlength{\tabcolsep}{2.5pt}
    \renewcommand{\arraystretch}{1.08}
    \resizebox{\textwidth}{!}{%
    
    \begin{threeparttable}
        \caption{Performance comparison on FB-100, FB-75, FB-50, and FB-25 under the fully inductive setting. The best and second-best results are highlighted in bold and underlined, respectively.}
        \label{tab:fully-inductive-fb}

        \begin{tabular}{@{}l*{12}{c}@{}}
            \toprule
            \multirow{2}{*}{\textbf{Model}}
            & \multicolumn{3}{c}{\textbf{FB-100}}
            & \multicolumn{3}{c}{\textbf{FB-75}}
            & \multicolumn{3}{c}{\textbf{FB-50}}
            & \multicolumn{3}{c}{\textbf{FB-25}} \\
            \cmidrule(lr){2-4}\cmidrule(lr){5-7}
            \cmidrule(lr){8-10}\cmidrule(lr){11-13}
            & MRR & H@1 & H@10
            & MRR & H@1 & H@10
            & MRR & H@1 & H@10
            & MRR & H@1 & H@10 \\
            \midrule
            CompGCN\tnote{*} & 0.015 & 0.008 & 0.025 & 0.013 & 0.000 & 0.026 & 0.004 & 0.002 & 0.006 & 0.003 & 0.000 & 0.004 \\
            NodePiece\tnote{*} & 0.006 & 0.001 & 0.009 & 0.016 & 0.007 & 0.029 & 0.021 & 0.006 & 0.048 & 0.044 & 0.011 & 0.114 \\
            NeuralLP\tnote{*} & 0.026 & 0.007 & 0.057 & 0.056 & 0.030 & 0.099 & 0.088 & 0.043 & 0.184 & 0.164 & 0.098 & 0.309 \\
            DRUM\tnote{*} & 0.034 & 0.011 & 0.077 & 0.065 & 0.034 & 0.121 & 0.101 & 0.061 & 0.191 & 0.175 & 0.109 & 0.320 \\
            BLP\tnote{*} & 0.017 & 0.004 & 0.035 & 0.047 & 0.024 & 0.085 & 0.078 & 0.037 & 0.156 & 0.107 & 0.053 & 0.212 \\
            QBLP\tnote{*} & 0.013 & 0.003 & 0.026 & 0.041 & 0.017 & 0.084 & 0.071 & 0.030 & 0.147 & 0.104 & 0.043 & 0.226 \\
            NBFNet\tnote{*} & 0.072 & 0.026 & 0.154 & 0.089 & 0.048 & 0.166 & 0.130 & 0.071 & 0.259 & 0.224 & 0.137 & 0.410 \\
            RED-GNN\tnote{*} & 0.121 & 0.053 & 0.263 & 0.107 & 0.057 & 0.201 & 0.129 & 0.072 & 0.251 & 0.145 & 0.077 & 0.284 \\
            RAILD\tnote{*} & 0.031 & 0.016 & 0.048 & \textendash & \textendash & \textendash & \textendash & \textendash & \textendash & \textendash & \textendash & \textendash \\
            INGRAM\tnote{*} & 0.223 & 0.146 & 0.371 & 0.189 & 0.119 & 0.325 & 0.117 & 0.067 & 0.218 & 0.133 & 0.067 & 0.271 \\
            Context Pooling\tnote{\#} & 0.146 & 0.056 & 0.323 & 0.128 & 0.074 & 0.242 & 0.144 & 0.069 & 0.307 & 0.235 & 0.142 & 0.425 \\
            TARGI\tnote{*} & 0.253 & 0.172 & 0.429 & \textendash & \textendash & \textendash & 0.159 & 0.088 & 0.307 & \textendash & \textendash & \textendash \\
            GraphOracle\tnote{\textasciicircum} & \underline{0.377} & \underline{0.277} & \underline{0.573} & \underline{0.315} & \underline{0.215} & \underline{0.510} & \underline{0.190} & \underline{0.107} & \underline{0.365} & \underline{0.276} & \underline{0.164} & \underline{0.513} \\
            \midrule
            QORH & \textbf{0.456} & \textbf{0.358} & \textbf{0.651} & \textbf{0.401} & \textbf{0.295} & \textbf{0.597} & \textbf{0.322} & \textbf{0.224} & \textbf{0.522} & \textbf{0.376} & \textbf{0.255} & \textbf{0.625} \\
            \bottomrule
        \end{tabular}%
        
        \begin{tablenotes}[flushleft]
            \footnotesize
            \item[*] Results are taken directly from \cite{leeINGRAMInductiveKnowledge2023}, \cite{duGraphOracleEfficientFullyInductive2026a}, or \cite{dingGlobalTopologyRelationGraph2025}.
            \item[\#] Results are reproduced using the official code under the
            training and evaluation protocols for the fully inductive setting.
            \item[\textendash] The corresponding result is not available.
            \item[\textasciicircum] Results are reproduced using the official
            code, with the data loading, training, and evaluation procedures
            adapted to the fully inductive training and evaluation protocols.
            See the \emph{Baselines and Experiment Setup} section for details.
        \end{tablenotes}
    \end{threeparttable}
    }
\end{table*}

\begin{table*}[t]
    \centering
    \small
    \setlength{\tabcolsep}{2.5pt}
    \renewcommand{\arraystretch}{1.08}
    \resizebox{\textwidth}{!}{%
    \begin{threeparttable}
        \caption{Performance comparison on NELL-100, NELL-75,
        NELL-50, and NELL-25 under the fully inductive setting.
        The best and second-best results are highlighted in bold
        and underlined, respectively.}
        \label{tab:fully-inductive-nell}
        \begin{tabular}{@{}l*{12}{c}@{}}
            \toprule
            \multirow{2}{*}{\textbf{Model}}
            & \multicolumn{3}{c}{\textbf{NL-100}}
            & \multicolumn{3}{c}{\textbf{NL-75}}
            & \multicolumn{3}{c}{\textbf{NL-50}}
            & \multicolumn{3}{c}{\textbf{NL-25}} \\
            \cmidrule(lr){2-4}\cmidrule(lr){5-7}
            \cmidrule(lr){8-10}\cmidrule(lr){11-13}
            & MRR & H@1 & H@10
            & MRR & H@1 & H@10
            & MRR & H@1 & H@10
            & MRR & H@1 & H@10 \\
            \midrule
            GraIL\tnote{*}
            & 0.135 & 0.114 & 0.173 & 0.096 & 0.056 & 0.205
            & 0.162 & 0.104 & 0.288 & 0.216 & 0.160 & 0.366 \\
            CoMPILE\tnote{*}
            & 0.123 & 0.071 & 0.209 & 0.178 & 0.093 & 0.361
            & 0.194 & 0.125 & 0.330 & 0.189 & 0.115 & 0.324 \\
            SNRI\tnote{*}
            & 0.042 & 0.029 & 0.064 & 0.088 & 0.040 & 0.177
            & 0.130 & 0.095 & 0.187 & 0.190 & 0.140 & 0.270 \\
            INDIGO\tnote{*}
            & 0.160 & 0.109 & 0.247 & 0.121 & 0.098 & 0.156
            & 0.167 & 0.134 & 0.217 & 0.166 & 0.134 & 0.206 \\
            RMPI\tnote{*}
            & 0.220 & 0.136 & 0.376 & 0.138 & 0.061 & 0.275
            & 0.185 & 0.109 & 0.307 & 0.213 & 0.130 & 0.329 \\
            CompGCN\tnote{*}
            & 0.008 & 0.001 & 0.014 & 0.014 & 0.003 & 0.025
            & 0.003 & 0.000 & 0.005 & 0.006 & 0.000 & 0.010 \\
            NodePiece\tnote{*}
            & 0.012 & 0.004 & 0.018 & 0.042 & 0.020 & 0.081
            & 0.037 & 0.013 & 0.079 & 0.098 & 0.057 & 0.166 \\
            NeuralLP\tnote{*}
            & 0.084 & 0.035 & 0.181 & 0.117 & 0.048 & 0.273
            & 0.101 & 0.064 & 0.190 & 0.148 & 0.101 & 0.271 \\
            DRUM\tnote{*}
            & 0.076 & 0.044 & 0.138 & 0.152 & 0.072 & 0.313
            & 0.107 & 0.070 & 0.193 & 0.161 & 0.119 & 0.264 \\
            BLP\tnote{*}
            & 0.019 & 0.006 & 0.037 & 0.051 & 0.012 & 0.120
            & 0.041 & 0.011 & 0.093 & 0.049 & 0.024 & 0.095 \\
            QBLP\tnote{*}
            & 0.004 & 0.000 & 0.003 & 0.040 & 0.007 & 0.095
            & 0.048 & 0.020 & 0.097 & 0.073 & 0.027 & 0.151 \\
            NBFNet\tnote{*}
            & 0.096 & 0.032 & 0.199 & 0.137 & 0.077 & 0.255
            & 0.225 & 0.161 & 0.346 & 0.283 & 0.224 & 0.417 \\
            RED-GNN\tnote{*}
            & 0.212 & 0.114 & 0.385 & 0.203 & 0.129 & 0.353
            & 0.179 & 0.115 & 0.280 & 0.214 & 0.166 & 0.266 \\
            RAILD\tnote{*}
            & 0.018 & 0.005 & 0.037
            & \textendash & \textendash & \textendash
            & \textendash & \textendash & \textendash
            & \textendash & \textendash & \textendash \\
            INGRAM\tnote{*}
            & 0.309 & 0.212 & \underline{0.506}
            & \underline{0.261} & \underline{0.167} & \underline{0.464}
            & 0.281 & 0.193 & 0.453
            & \underline{0.334} & \underline{0.241} & \textbf{0.501} \\
            Context Pooling\tnote{\#}
            & 0.219 & 0.138 & 0.370 & 0.180 & 0.099 & 0.336
            & 0.226 & 0.146 & 0.363 & 0.231 & 0.170 & 0.307 \\
            TARGI\tnote{*}
            & \underline{0.314} & \underline{0.230} & 0.489
            & \textendash & \textendash & \textendash
            & \underline{0.314} & \underline{0.207} & \textbf{0.533}
            & \textendash & \textendash & \textendash \\
            GraphOracle\tnote{\textasciicircum}
            & 0.247 & 0.149 & 0.435 & 0.244 & 0.155 & 0.424
            & 0.255 & 0.169 & 0.420 & 0.261 & 0.173 & 0.400 \\
            \midrule
            QORH
            & \textbf{0.462} & \textbf{0.363} & \textbf{0.644}
            & \textbf{0.336} & \textbf{0.245} & \textbf{0.536}
            & \textbf{0.380} & \textbf{0.303} & \underline{0.519}
            & \textbf{0.365} & \textbf{0.294} & \underline{0.475} \\
            \bottomrule
        \end{tabular}%
        \begin{tablenotes}[flushleft]
            \footnotesize
            \item[*] Results are taken directly from \cite{leeINGRAMInductiveKnowledge2023}, \cite{duGraphOracleEfficientFullyInductive2026a}, or \cite{dingGlobalTopologyRelationGraph2025}.
            \item[\#] Results are reproduced using the official code
            under the training and evaluation protocols for the
            fully inductive setting.
            \item[\textendash] The corresponding result is not available.
            \item[\textasciicircum] Results are reproduced using the
            official code, with the data loading, training, and
            evaluation procedures adapted to the fully inductive
            training and evaluation protocols. See the
            \emph{Baselines and Experiment Setup} section for details.
        \end{tablenotes}
    \end{threeparttable}%
    }
\end{table*}

\begin{table*}[t]
    \centering
    \small
    \setlength{\tabcolsep}{2.5pt}
    \renewcommand{\arraystretch}{1.08}
    \resizebox{\textwidth}{!}{%
    \begin{threeparttable}
        \caption{Performance comparison on WK-100, WK-75,
        WK-50, and WK-25 under the fully inductive setting.
        The best and second-best results are highlighted in bold
        and underlined, respectively.}
        \label{tab:fully-inductive-wk}
        \begin{tabular}{@{}l*{12}{c}@{}}
            \toprule
            \multirow{2}{*}{\textbf{Model}}
            & \multicolumn{3}{c}{\textbf{WK-100}}
            & \multicolumn{3}{c}{\textbf{WK-75}}
            & \multicolumn{3}{c}{\textbf{WK-50}}
            & \multicolumn{3}{c}{\textbf{WK-25}} \\
            \cmidrule(lr){2-4}\cmidrule(lr){5-7}
            \cmidrule(lr){8-10}\cmidrule(lr){11-13}
            & MRR & H@1 & H@10
            & MRR & H@1 & H@10
            & MRR & H@1 & H@10
            & MRR & H@1 & H@10 \\
            \midrule
            CompGCN\tnote{*}
            & 0.003 & 0.000 & 0.009 & 0.015 & 0.003 & 0.028
            & 0.003 & 0.001 & 0.002 & 0.009 & 0.000 & 0.020 \\
            NodePiece\tnote{*}
            & 0.007 & 0.002 & 0.018 & 0.021 & 0.003 & 0.052
            & 0.008 & 0.002 & 0.013 & 0.053 & 0.019 & 0.122 \\
            NeuralLP\tnote{*}
            & 0.009 & 0.005 & 0.016 & 0.020 & 0.004 & 0.054
            & 0.025 & 0.007 & 0.054 & 0.068 & 0.046 & 0.104 \\
            DRUM\tnote{*}
            & 0.010 & 0.004 & 0.019 & 0.020 & 0.007 & 0.043
            & 0.017 & 0.002 & 0.046 & 0.064 & 0.035 & 0.116 \\
            BLP\tnote{*}
            & 0.012 & 0.003 & 0.025 & 0.043 & 0.016 & 0.089
            & 0.041 & 0.013 & 0.092 & 0.125 & 0.055 & 0.283 \\
            QBLP\tnote{*}
            & 0.012 & 0.003 & 0.025 & 0.044 & 0.016 & 0.091
            & 0.035 & 0.011 & 0.080 & 0.116 & 0.042 & 0.294 \\
            NBFNet\tnote{*}
            & 0.014 & 0.005 & 0.026 & 0.072 & 0.028 & 0.172
            & 0.062 & 0.036 & 0.105 & 0.154 & 0.092 & 0.301 \\
            RED-GNN\tnote{*}
            & 0.096 & 0.070 & 0.136 & 0.172 & 0.110 & 0.290
            & 0.058 & 0.033 & 0.093 & 0.170 & 0.111 & 0.263 \\
            RAILD\tnote{*}
            & 0.026 & 0.010 & 0.052
            & \textendash & \textendash & \textendash
            & \textendash & \textendash & \textendash
            & \textendash & \textendash & \textendash \\
            INGRAM\tnote{*}
            & 0.107 & 0.072 & 0.169
            & \underline{0.247} & \underline{0.179} & 0.362
            & \underline{0.068} & 0.034 & \underline{0.135}
            & \underline{0.186} & \underline{0.124} & \underline{0.309} \\
            Context Pooling\tnote{\#}
            & 0.117 & 0.082 & 0.185 & 0.181 & 0.100 & 0.322
            & 0.032 & 0.011 & 0.065 & 0.151 & 0.078 & 0.282 \\
            GraphOracle\tnote{\textasciicircum}
            & \underline{0.139} & \underline{0.097} & \underline{0.231}
            & 0.204 & 0.123 & \underline{0.371}
            & 0.057 & \underline{0.036} & 0.096
            & 0.132 & 0.050 & 0.278 \\
            \midrule
            QORH
            & \textbf{0.181} & \textbf{0.128} & \textbf{0.282}
            & \textbf{0.375} & \textbf{0.291} & \textbf{0.513}
            & \textbf{0.135} & \textbf{0.080} & \textbf{0.243}
            & \textbf{0.276} & \textbf{0.195} & \textbf{0.429} \\
            \bottomrule
        \end{tabular}%
        \begin{tablenotes}[flushleft]
            \footnotesize
            \item[*] Results are taken directly from \cite{leeINGRAMInductiveKnowledge2023} or \cite{duGraphOracleEfficientFullyInductive2026a}.
            \item[\#] Results are reproduced using the official code
            under the training and evaluation protocols for the
            fully inductive setting.
            \item[\textendash] The corresponding result is not available.
            \item[\textasciicircum] Results are reproduced using the
            official code, with the data loading, training, and
            evaluation procedures adapted to the fully inductive
            training and evaluation protocols. See the
            \emph{Baselines and Experiment Setup} section for details.
        \end{tablenotes}
    \end{threeparttable}%
    }
\end{table*}

Tables~\ref{tab:entity-inductive-results}, \ref{tab:fully-inductive-fb}, \ref{tab:fully-inductive-nell}, and \ref{tab:fully-inductive-wk} report the comparative results of QORH and the baselines on entity-inductive and fully inductive link prediction tasks. As shown in Table~\ref{tab:entity-inductive-results}, QORH achieves the highest MRR and Hits@1 on 11 of the 12 entity-inductive datasets and the highest Hits@10 on all 12 datasets. For example, on FB15k-237-v1, QORH improves MRR from 0.383, achieved by Context Pooling~\cite{suContextPoolingQueryspecific2025}, to 0.466, corresponding to a relative improvement of 21.67\%. This result demonstrates its performance advantage over existing query-aware graph reasoning methods.

Tables~\ref{tab:fully-inductive-fb}, \ref{tab:fully-inductive-nell}, \ref{tab:fully-inductive-wk} present the fully inductive results for the FB, NL, and WK families, respectively. QORH achieves the highest MRR and Hits@1 on all 12 fully inductive datasets and the highest Hits@10 on 10 of them. In particular, on FB-100, NL-100, and WK-100, both the entity sets and the relation sets are completely disjoint between the training and inference graphs. This entity- and relation-disjoint setting requires the model to generalize simultaneously to unseen entities and relations, making it more challenging. On these three datasets, QORH achieves MRR scores of 0.456, 0.462, and 0.181, respectively, outperforming the best baseline on each dataset by 20.95\%, 47.13\%, and 30.22\%. These results demonstrate that QORH can effectively exploit structural information in the inference graph to predict missing links even when no entities or relations are shared between training and inference, further supporting its inductive generalization capability under simultaneous changes in entities and relations.

\subsection{Ablation Studies}
\begin{table*}[t]
    \centering
    \small
    \setlength{\tabcolsep}{2.5pt}
    \renewcommand{\arraystretch}{1.08}
    \resizebox{\textwidth}{!}{%
    \begin{threeparttable}
        \caption{Ablation results on the v1--v4 splits of WN18RR,
        FB15k-237, and NELL-995 under the entity-inductive setting.
        The best results are highlighted in bold.}
        \label{tab:qorh-ablation}
        \begin{tabular}{@{}ll*{9}{c}@{}}
            \toprule
            \multirow{2}{*}{\textbf{Dataset}}
            & \multirow{2}{*}{\textbf{Version}}
            & \multicolumn{3}{c}{\textbf{QORH}}
            & \multicolumn{3}{c}{\textbf{w/o RHE}}
            & \multicolumn{3}{c}{\textbf{w/o RS}} \\
            \cmidrule(lr){3-5}\cmidrule(lr){6-8}
            \cmidrule(lr){9-11}
            & & MRR & H@1 & H@10
            & MRR & H@1 & H@10
            & MRR & H@1 & H@10 \\
            \midrule

            \multirow{4}{*}{WN18RR} & v1
            & \textbf{0.793} & \textbf{0.753} & \textbf{0.878}
            & 0.737 & 0.681 & 0.843
            & 0.738 & 0.665 & 0.859 \\
            & v2
            & \textbf{0.725} & \textbf{0.678} & \textbf{0.801}
            & 0.676 & 0.602 & 0.793
            & 0.693 & 0.629 & 0.801 \\
            & v3
            & \textbf{0.490} & \textbf{0.423} & \textbf{0.603}
            & 0.453 & 0.385 & 0.569
            & 0.454 & 0.395 & 0.564 \\
            & v4
            & \textbf{0.679} & \textbf{0.638} & \textbf{0.752}
            & 0.128 & 0.057 & 0.280
            & 0.627 & 0.560 & 0.735 \\
            \midrule

            \multirow{4}{*}{FB15k-237} & v1
            & \textbf{0.466} & \textbf{0.381} & \textbf{0.600}
            & 0.454 & 0.373 & 0.573
            & 0.454 & 0.378 & 0.571 \\
            & v2
            & \textbf{0.533} & \textbf{0.443} & \textbf{0.694}
            & 0.508 & 0.409 & 0.686
            & 0.499 & 0.384 & 0.694 \\
            & v3
            & \textbf{0.517} & \textbf{0.425} & \textbf{0.677}
            & 0.495 & 0.395 & 0.663
            & 0.500 & 0.402 & 0.660 \\
            & v4
            & \textbf{0.497} & \textbf{0.399} & \textbf{0.676}
            & 0.260 & 0.154 & 0.483
            & 0.478 & 0.375 & 0.670 \\
            \midrule

            \multirow{4}{*}{NELL-995} & v1
            & 0.649 & \textbf{0.550} & \textbf{0.950}
            & 0.573 & 0.505 & 0.780
            & \textbf{0.679} & 0.535 & 0.945 \\
            & v2
            & \textbf{0.611} & \textbf{0.507} & \textbf{0.792}
            & 0.489 & 0.374 & 0.704
            & 0.597 & 0.498 & 0.766 \\
            & v3
            & \textbf{0.567} & \textbf{0.466} & \textbf{0.745}
            & 0.546 & 0.453 & 0.712
            & 0.549 & 0.454 & 0.706 \\
            & v4
            & \textbf{0.535} & \textbf{0.415} & \textbf{0.741}
            & 0.243 & 0.186 & 0.350
            & 0.358 & 0.266 & 0.546 \\
            \midrule

            \multicolumn{2}{@{}l}{\textbf{Macro-average}}
            & \textbf{0.588} & \textbf{0.506} & \textbf{0.742}
            & 0.463 & 0.381 & 0.620
            & 0.552 & 0.462 & 0.710 \\
            \bottomrule
        \end{tabular}
        \begin{tablenotes}[flushleft]
            \footnotesize
            \item \textit{Notes.} w/o RHE and w/o RS denote QORH variants without rule hypergraph enhancement and learnable rule selection, respectively. H@1 and H@10 denote Hits@1 and Hits@10. Macro-average is the unweighted mean over the 12 dataset splits. For each split and metric, the best result across the three variants is shown in bold.
        \end{tablenotes}
    \end{threeparttable}%
    }
\end{table*}

To investigate the contributions of ordered rule hypergraph enhancement and rule selection, we conduct ablation studies on the 12 entity-inductive datasets using two variants. QORH w/o RHE removes the entire rule hypergraph enhancement module while retaining the base relation encoding and entity-level reasoning. QORH w/o RS removes the rule selection module and uses all candidate rules for hypergraph enhancement while retaining ordered rule encoding and hypergraph message passing. Detailed results are reported in Table~\ref{tab:qorh-ablation}, where the averages are computed as the unweighted means across the 12 dataset variants.

Removing rule hypergraph enhancement degrades all three metrics on all 12 datasets. Average MRR, Hits@1, and Hits@10 decrease from 0.5885, 0.5064, and 0.7423 to 0.4634, 0.3812, and 0.6195, respectively, with average MRR dropping by 21.25\% relative to the full model. These results demonstrate the effectiveness of rule hypergraph enhancement for inductive reasoning. The base relation graph captures pairwise dependencies but omits higher-order joint information involving multiple relations.

Removing rule selection reduces average MRR, Hits@1, and Hits@10 to 0.5521, 0.4617, and 0.7095, corresponding to relative decreases of 6.17\%, 8.83\%, and 4.41\% compared with the full model. These results indicate that directly incorporating all candidate rules does not yield consistent performance gains. This trend is consistent with the rationale for selecting rules based on confidence, coverage, rule length, and support, indicating that rule selection improves the usefulness of the incorporated evidence.

The ablation results show that rule hypergraph enhancement and rule selection serve distinct roles: the ordered rule hypergraph provides information complementary to the base relation encoding, while rule selection further improves the utilization of this evidence.

\subsection{Hyperparameter Sensitivity Analysis}

To assess the sensitivity of the ordered rule hypergraph module to its key hyperparameters, we examine the maximum number of rules per query relation K and the maximum rule-body length $L_{\max}$ in both entity-inductive and fully inductive settings. With $K\in\{1,2,3,4,5,6,7\}$ and $L_{\max}\in\{2,3\}$, we compare 14 parameter combinations on each dataset.

\begin{figure}[htbp]
  \centering
  \includegraphics[width=1.0\textwidth]{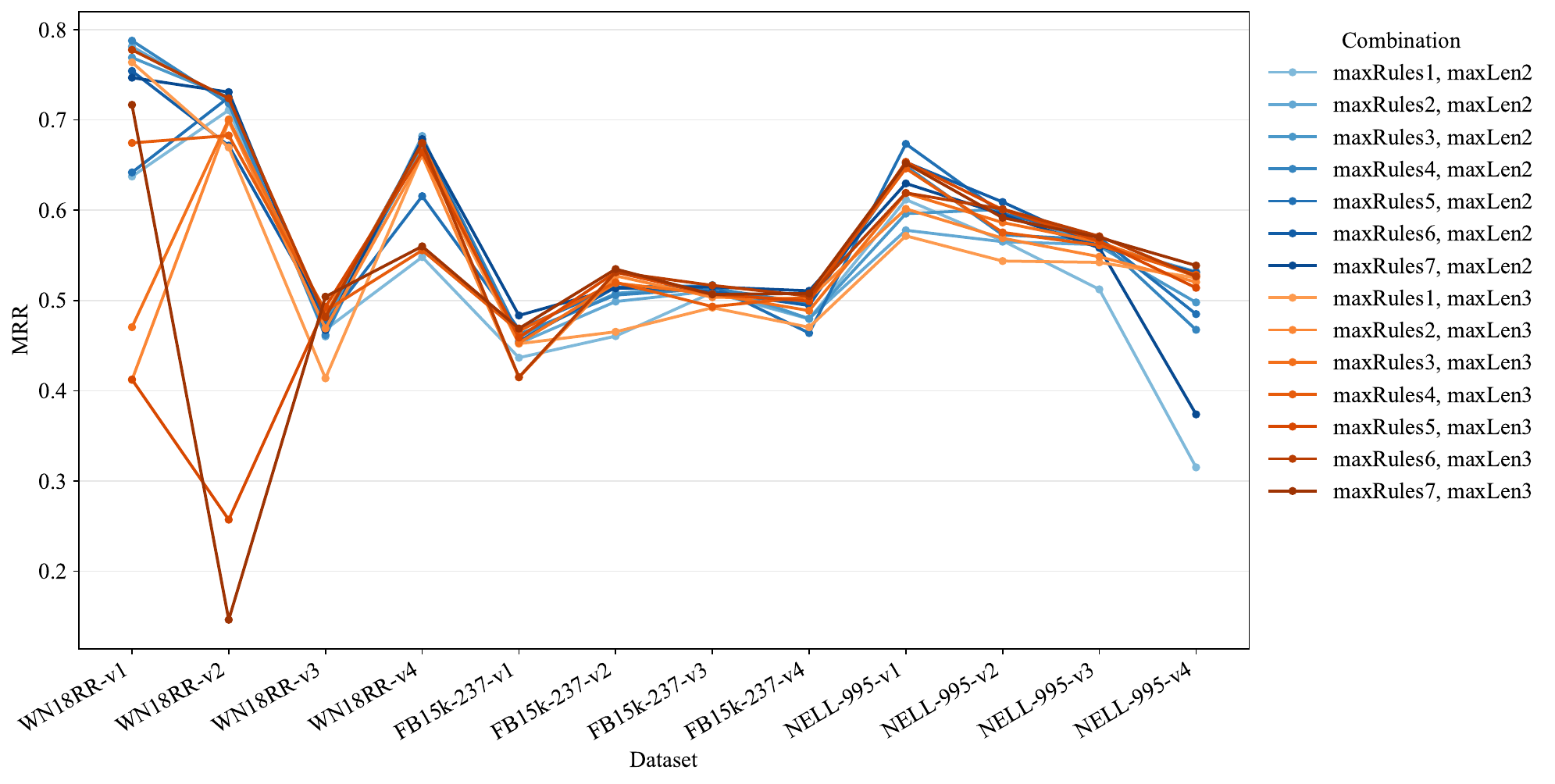}
  \caption{MRR of hyperparameter combination across entity-inductive datasets.}
  \label{fig:entity-inductive-hyperparameter1}
\end{figure}

\begin{figure}[htbp]
  \centering
  \includegraphics[width=1.0\textwidth]{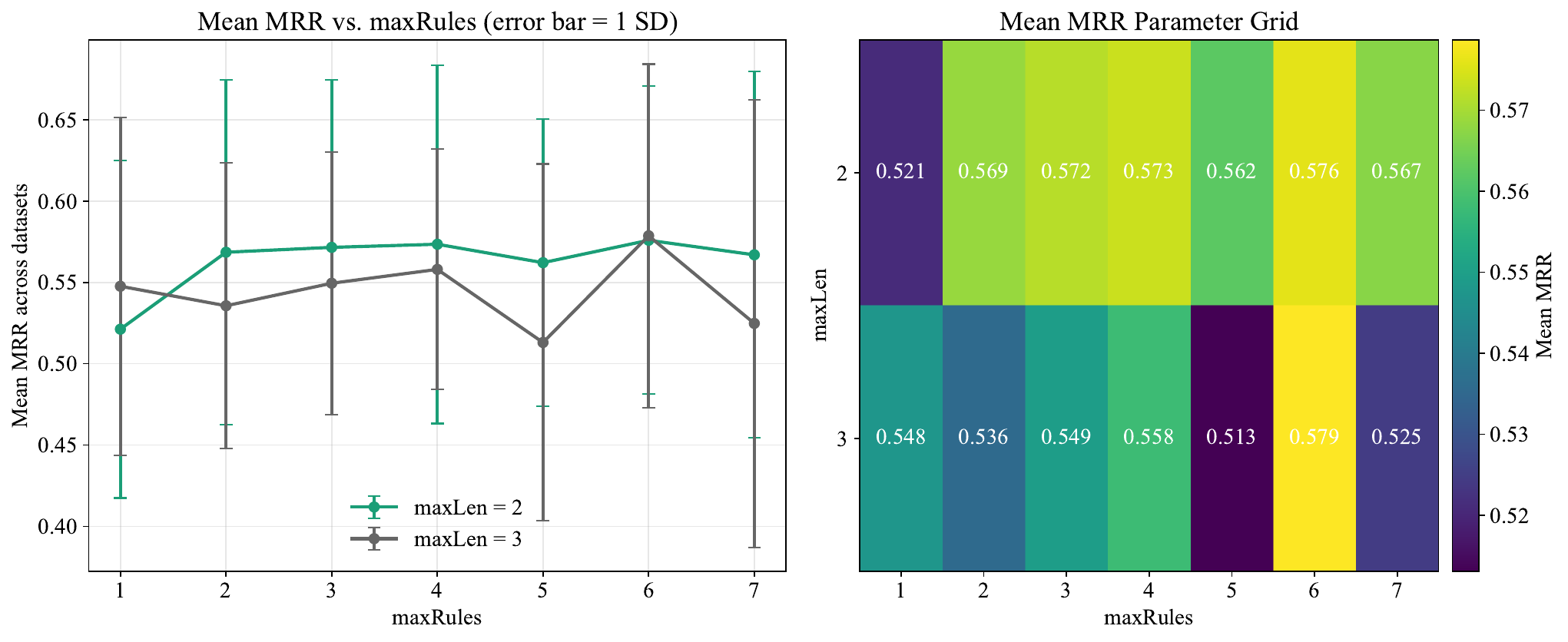}
  \caption{Hyperparameter sensitivity summary on entity-inductive Datasets.}
  \label{fig:entity-inductive-hyperparameter4}
\end{figure}

\begin{figure}[htbp]

  \centering
  \includegraphics[width=1.0\textwidth]{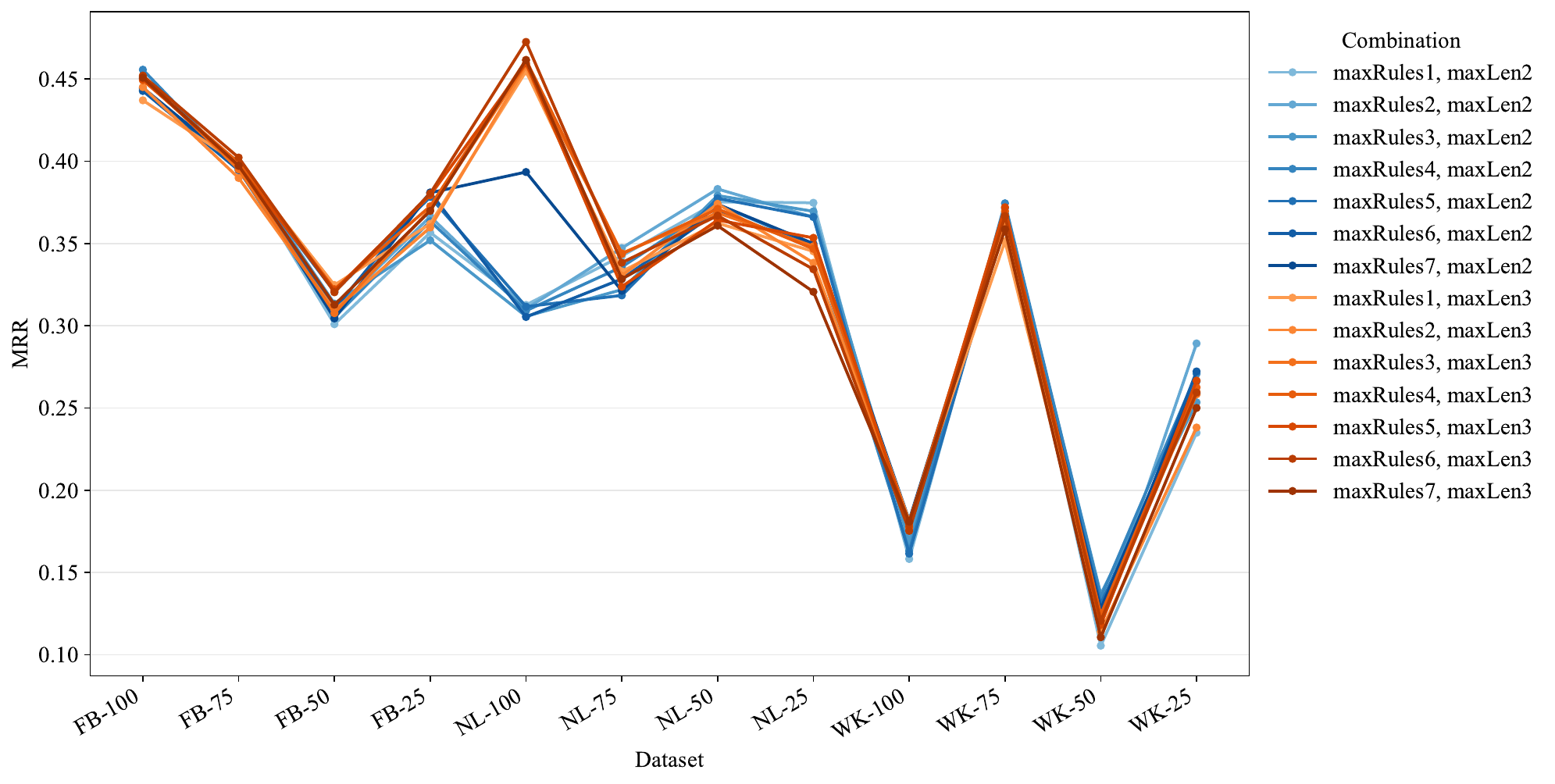}
  \caption{MRR of hyperparameter combination across fully inductive datasets.}
  \label{fig:fully-inductive-hyperparameter1}
\end{figure}

\begin{figure}[htbp]
  \centering
  \includegraphics[width=1.0\textwidth]{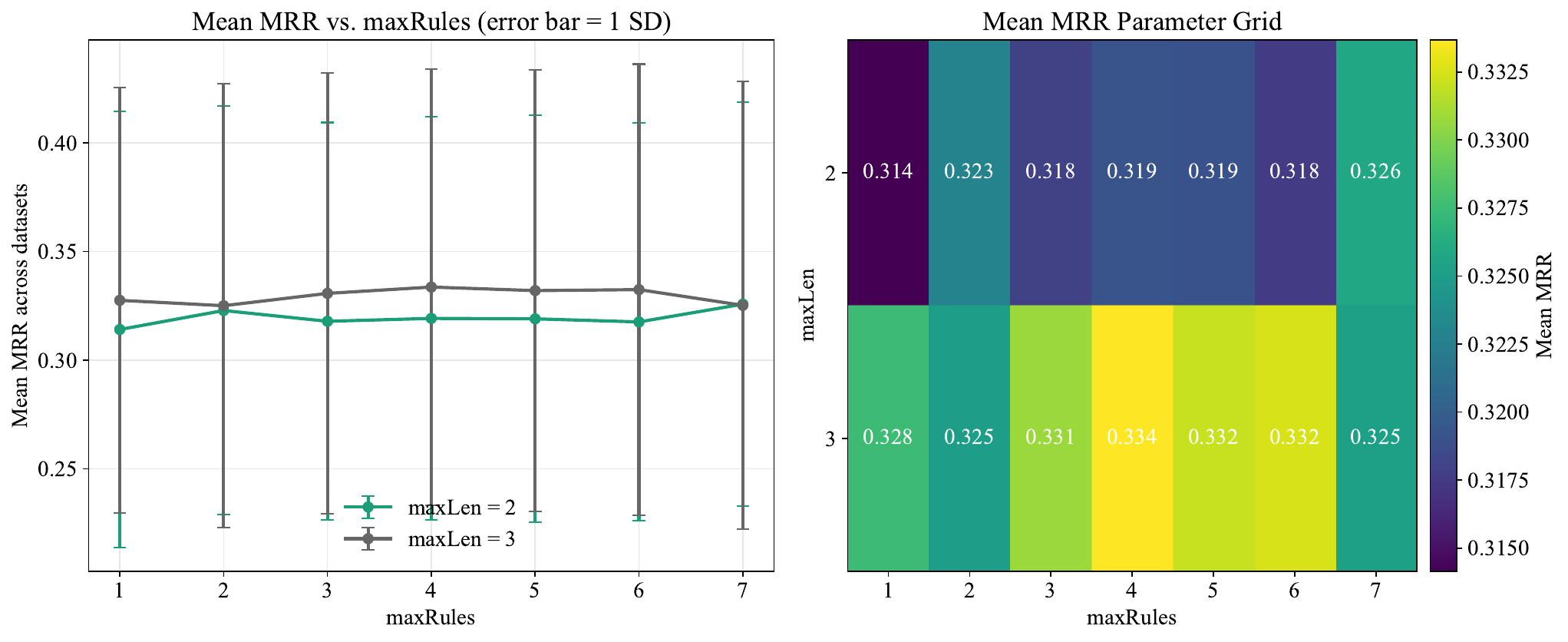}
  \caption{Hyperparameter sensitivity summary on fully inductive Datasets.}
  \label{fig:fully-inductive-hyperparameter4}
\end{figure}

The effect of rule length varies across datasets. As shown in Figure~\ref{fig:fully-inductive-hyperparameter1}, in the fully inductive setting, $L_{\max}=3$ yields modest improvements over $L_{\max}=2$ on FB-25, FB-50, and WK-100, with a more pronounced improvement on NL-100. As shown in Figure~\ref{fig:entity-inductive-hyperparameter1}, in the entity-inductive setting, $L_{\max}=3$ provides slight gains on WN18RR-v3 and FB15k-237-v2 and a more substantial gain on NELL-995-v4. Accordingly, we use $L_{\max}=3$ for these datasets and $L_{\max}=2$ for the remaining datasets.

As shown in Figures~\ref{fig:entity-inductive-hyperparameter4} and \ref{fig:fully-inductive-hyperparameter4}, performance varies only slightly across different values of $K$ in the fully inductive setting, indicating that the model is relatively insensitive to this hyperparameter. Across both inductive settings, $K=4$ provides consistently robust performance across datasets. We therefore set $K=4$ for all datasets. All QORH results reported in Section~\ref{sec:Main Results} use these hyperparameter configurations.

\section{Conclusin}

We propose QORH, a query-specific ordered rule hypergraph method, to address the difficulty of preserving complete rule composition information through pairwise relational dependencies in inductive knowledge graph link prediction. Conditioned on the query relation, QORH jointly encodes rule heads, ordered rule bodies, and statistical evidence as hyperedges. Through learnable rule selection and role-aware relation–hyperedge–relation propagation, it incorporates joint dependencies among multiple relations into relation representations, which subsequently guide entity-level reasoning. Extensive experiments in both entity-inductive and fully inductive settings show that QORH substantially outperforms existing state-of-the-art baselines, demonstrating its effectiveness and robustness. In future work, we will explore joint modeling of rule structures and textual semantics, investigate rule generation and selection assisted by large language models, and extend ordered rule hypergraphs to a broader range of structured reasoning tasks.

\bibliographystyle{splncs04_unsort} 
\bibliography{ref261013}

\end{document}